%% file: mlsys2025style/main.tex
\documentclass{article}
\usepackage{listings}
\usepackage{xcolor}
\usepackage{graphicx}
\usepackage{subcaption}
\definecolor{codegray}{rgb}{0.95,0.95,0.95}
\usepackage{amsmath}
\usepackage[utf8]{inputenc}

\usepackage{microtype}
\usepackage{graphicx}
\usepackage{subcaption}
\usepackage{booktabs}
\usepackage{hyperref}
\usepackage{booktabs}
\usepackage[most]{tcolorbox}
\usepackage{tabularx}
\usepackage{multirow}
\usepackage[table]{xcolor}
\usepackage{makecell}  
\usepackage{multirow}
\usepackage[table]{xcolor}
\usepackage{makecell}
\usepackage{float}

\definecolor{RowBlue}{HTML}{ECF4FF}
\definecolor{RowGreen}{HTML}{E8F7EC}
\definecolor{RowGray}{HTML}{F4F4F4}
\definecolor{CellYellow}{HTML}{FFFFE0} 
\usepackage{xspace} 
\definecolor{RowBlue}{HTML}{ECF4FF}
\definecolor{RowGreen}{HTML}{E8F7EC}
\definecolor{RowGray}{HTML}{F4F4F4}

\usepackage{amsmath}
\usepackage{amsfonts}
\usepackage{amssymb}
\usepackage{float}
\usepackage[htt]{hyphenat} 
\usepackage{array}     
\usepackage{lineno}

\usepackage{url}
\usepackage{balance}

\usepackage{tikz}
\usepackage{etoolbox}
\usepackage{algorithm}
\usepackage{algpseudocode}
\usepackage{xcolor}
\usepackage{calc}
\usepackage{mlsys2025}
\mlsystitlerunning{}
\usepackage{textcomp}

\lstdefinestyle{cppstyle}{
    language=C++,
    basicstyle=\ttfamily\footnotesize,
    numbers=left,
    numberstyle=\tiny,
    stepnumber=1,
    numbersep=6pt,
    breaklines=true,
    breakatwhitespace=false,
    showstringspaces=false,
    tabsize=2,
    frame=none,
    keywordstyle=\color{blue},
    commentstyle=\color{gray},
    stringstyle=\color{teal}
}

\newcommand{\highlightyellow}[1]{%
  \leavevmode
  \setlength{\fboxsep}{0pt}%
  \colorbox{yellow!20}{\hbox to \dimexpr\linewidth-\leftmargin\relax{\strut #1\hfill}}%
}
\newcommand{\highlightblue}[1]{%
  \leavevmode
  \setlength{\fboxsep}{0pt}%
  \colorbox{blue!10}{\hbox to \dimexpr\linewidth-\leftmargin\relax{\strut #1\hfill}}%
}
\newcommand{\KB}{\textit{Knowledge Base}\xspace}
\newcommand{\sys}{\textsc{KernelBlaster}\xspace}
\makeatletter
\renewcommand{\@listI}{%
  \leftmargin\leftmargini
  \parsep 0pt
  \topsep 2pt 
  \itemsep 1pt 
  \partopsep 0pt
}

\renewcommand{\@listii}{%
  \leftmargin\leftmarginii
  \labelwidth\leftmarginii
  \advance\labelwidth-\labelsep
  \topsep 1pt
  \parsep 0pt
  \itemsep 1pt
  \partopsep 0pt
}
\makeatother

\makeatletter
\long\def\@makecaption#1#2{
  \vskip -2pt  
  \baselineskip 1pt
  \setbox\@tempboxa\hbox{#1. #2}
  \ifdim \wd\@tempboxa >\hsize
    \sbox{\newcaptionbox}{\small\sl #1.~}
    \newcaptionboxwid=\wd\newcaptionbox
    \usebox\newcaptionbox {\footnotesize #2}
  \else
    \centerline{{\small\sl #1.} {\small #2}}
  \fi
  \vskip -10pt 

}
\makeatother

\makeatletter
\long\def\@makecaption#1#2{%
  \vspace{-1pt}\vskip 1pt 
  \setbox\@tempboxa\hbox{{\small\sl #1.} {\small #2}}%
  \ifdim \wd\@tempboxa >\hsize
    {\small\sl #1.} {\small #2}\par
  \else
    \centerline{{\small\sl #1.} {\small #2}}%
  \fi
  \vskip 1pt 
}
\makeatother

\begin{document}

\makeatletter
\AtBeginDocument{%
  \setlength{\footskip}{18pt} 
  \pagestyle{fancy}
  \fancyhf{}%
  \fancyfoot[C]{\thepage}%
  \renewcommand{\headrulewidth}{1pt}%
}
\makeatother

\twocolumn[
\mlsystitle{\sys: Continual Cross-Task CUDA Optimization via Memory-Augmented In-Context Reinforcement Learning}
\centering{Kris Shengjun Dong$^{1,2,\ast}$, Sahil Modi$^{1}$, Dima Nikiforov$^{2}$, Sana Damani$^{1}$,\\
Edward Lin$^{1}$, Siva Kumar Sastry Hari$^{1}$, Christos Kozyrakis$^{1}$\\
$^{1}$NVIDIA, $^{2}$University of California, Berkeley}


\mlsysaffiliation{anon}{Nivida, CA, USA}
\mlsyscorrespondingauthor{Kris Shengjun Dong}{krisd@nvidia.com}

\mlsyskeywords{Large-Language Model, In-context Reinforcement Learning, GPU Code Optimization}
\nolinenumbers

\vskip 0.3in
\begin{abstract}
\vspace{7pt}
Optimizing CUDA code across multiple generations of GPU architectures is challenging, as achieving peak performance requires an extensive exploration of an increasingly complex, hardware-specific optimization space. Traditional compilers are constrained by fixed heuristics, whereas finetuning Large Language Models (LLMs) can be expensive. However, agentic workflows for CUDA code optimization have limited ability to aggregate knowledge from prior exploration, leading to biased sampling and suboptimal solutions. We propose \textbf{\sys}, a Memory-Augmented In-context Reinforcement Learning (MAIC-RL) framework designed to improve CUDA optimization search capabilities of LLM-based GPU coding agents. {\sys} enables agents to learn from experience and make systematically informed decisions on future tasks by accumulating knowledge into a retrievable \textit{Persistent CUDA Knowledge Base}. We propose a novel profile-guided, textual-gradient-based agentic flow for CUDA generation and optimization to achieve high performance across generations of GPU architectures. {\sys} guides LLM agents to systematically explore high-potential optimization strategies beyond naive rewrites. Compared to the PyTorch baseline, our method achieves geometric mean speedups of 1.43$\times$, 2.50$\times$, and 1.50$\times$ on KernelBench Levels~1, 2, and 3, respectively. We release {\sys} as an open-source agentic framework, accompanied by a test harness, verification components, and a reproducible evaluation pipeline$^{\dagger}$.

\end{abstract}
]
\begingroup
  \renewcommand\thefootnote{\fnsymbol{footnote}}
  \footnotetext[1]{Most of this work was done by Kris Shengjun Dong during her 2025 summer internship at NVIDIA.}
\endgroup


\input{mlsys2025style/01_introduction}

\input{mlsys2025style/02_related_work}

\input{mlsys2025style/03_methodology}

\input{mlsys2025style/05_evaluation}
\input{mlsys2025style/06_AB_experiment}

\input{mlsys2025style/07_conclusion}

\bibliographystyle{mlsys2025}

\bibliography{references}

\newpage
\input{mlsys2025style/appendix}

\end{document}

%% file: mlsys2025style/01_introduction.tex
\section{Introduction}

As machine learning workloads evolve, emerging model architectures and increasingly dynamic workloads introduce new execution patterns that invalidate previously tuned kernels, posing new challenges to existing software stacks \cite{sevilla2022compute}. To sustain high performance under this shifting landscape, systems must continually adapt their optimization strategies. Achieving state-of-the-art efficiency increasingly depends on maintaining and extending specialized kernel libraries tuned for new operator patterns and hardware targets. This process has traditionally required substantial engineering effort and deep domain expertise. As frontier models rapidly evolve, this optimization process becomes a scalability bottleneck, preventing full usage of emerging hardware capabilities and limiting achievable performance\cite{sevilla2022compute}.

The rapid evolution of hardware exposes the limitations of manual kernel implementation tuning. One example is FlashAttention: when FlashAttention-2 \cite{dao2023flashattention2} was initially ported to NVIDIA’s H100 GPUs \cite{nvidia2022h100}, performance dropped by roughly 47\%, reflecting mismatches between prior optimizations and the new architecture. Only after a redesign effort did FlashAttention-3 \cite{shah2024flashattention3} introduce architecture-aware optimizations that recovered and exceeded prior efficiency. Another example is DeepSeek-V3, which particularly targets NVIDIA H800 GPUs \cite{nvidiaH800},  with training pipelines and communication schedules designed around that platform’s execution model, including FP8 computation and cluster-level data movement~\cite{deepseekv3}. These optimizations assume specific hardware characteristics and cluster topology, and transferring them to other GPU variants requires them to be redesigned. As hardware and workload characteristics evolve, platform-specific tuning can quickly lose relevance, motivating optimization approaches that adapt across hardware generations instead of relying on fixed, device-specific engineering.

This manual optimization pipeline increasingly becomes a scalability bottleneck, limiting how quickly software can exploit new hardware capabilities and pushing against practical performance ceilings. As a consequence, a research domain has recently emerged to investigate LLMs' capabilities for GPU code generation and optimization \cite{gim2025pie, lin2025eco, openevolve2025, Baronio2025Kevin32B, Cummins2025LLMCompiler, Damani2024Warpdrive, Taneja2025LlmVectorizer}. Initial studies have highlighted the potential for using LLMs to enhance GPU program performance \cite{peng2025sysllmatic, gong2025language, Lange2025AICUDAEngineer}. However, there remain substantial opportunities to improve generalizability, learning capability, sample efficiency, and cost when applying LLMs code optimization.

\begin{figure}[!t]
\centering
\includegraphics[width=\linewidth]{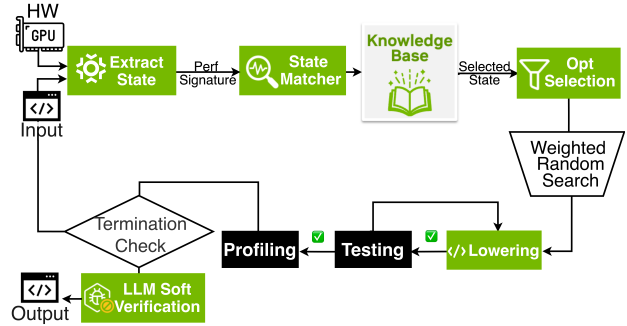}
\caption{High-level block diagram of the {\sys} agentic workflow.}
\label{fig:highlevelblock_diagram}
\end{figure}

We present \textbf{{\sys}}, a novel Memory-Augmented In-context Reinforcement Learning
(MAIC-RL) framework designed to automate the task of CUDA code optimization. {\sys} leverages ICRL techniques to build a \textit{Persistent CUDA Knowledge Base} (``\KB'') to allow LLM agents to learn and apply code-optimization policies from experience. ~\autoref{fig:highlevelblock_diagram} shows the high-level agentic system.

Our primary contributions are:
\begin{enumerate}
    \item \textbf{In-Context Reinforcement Learning (ICRL) Framework for CUDA Optimizations:} We formulate the CUDA code optimization problem as a reinforcement learning problem with textual gradient updates, capturing rich semantic information from profile data and enabling inference-time learning, enabling faster and more directed learning compared to prior RL methodologies that directly update model weights.
    \item \textbf{A Comprehensive \textit{Knowledge Base}}: We aggregate experience from past optimization attempts into a \textit{Knowledge Base} data structure, enabling efficient traversal of optimization candidates compared to prior static solutions. We propose a novel hierarchical representation that categorizes code instances into performance states, resulting in a scalable representation that efficiently utilizes model context.
    \item \textbf{A Framework for Long-Term Cross-Task Learning:} \sys simultaneously generates optimized kernels while also aggregating knowledge across problems, generating a re-usable artifact that enables faster convergence on new problems and GPU hardware platforms. This artifact is designed for adaptability and can be specialized for distinct application domains and specific GPU architectures (e.g., NVIDIA Ampere vs. Hopper), enabling the agent to apply accumulated experience effectively to future unseen problems.
    \item \textbf{An Open-Source Textual RL Framework}: We will release an open-source implementation of our textural RL agentic workflow, including baseline CUDA kernels, initialized databases, and test harnesses.\begingroup
  \renewcommand{\thefootnote}{\ensuremath{\dagger}}%
  \footnote{The repository will be released in a subsequent revision.}%
\endgroup

\end{enumerate}

Our experiments show that {\sys} guides LLM agents to systematically explore high-potential optimization strategies beyond naive rewrites across architectures, resulting in a geometric mean performance speedup of 1.43$\times$ over the PyTorch baseline on KernelBench Level 1, 2.50$\times$ on complex operator compositions in KernelBench Level 2 and 2.50$\times$ when accelerating entire models in KernelBench Level 3. 

%% file: mlsys2025style/02_related_work.tex
\section{Related Work}\label{sec:related}
\begin{figure*}[!h]
\centering
\includegraphics[width=\linewidth]{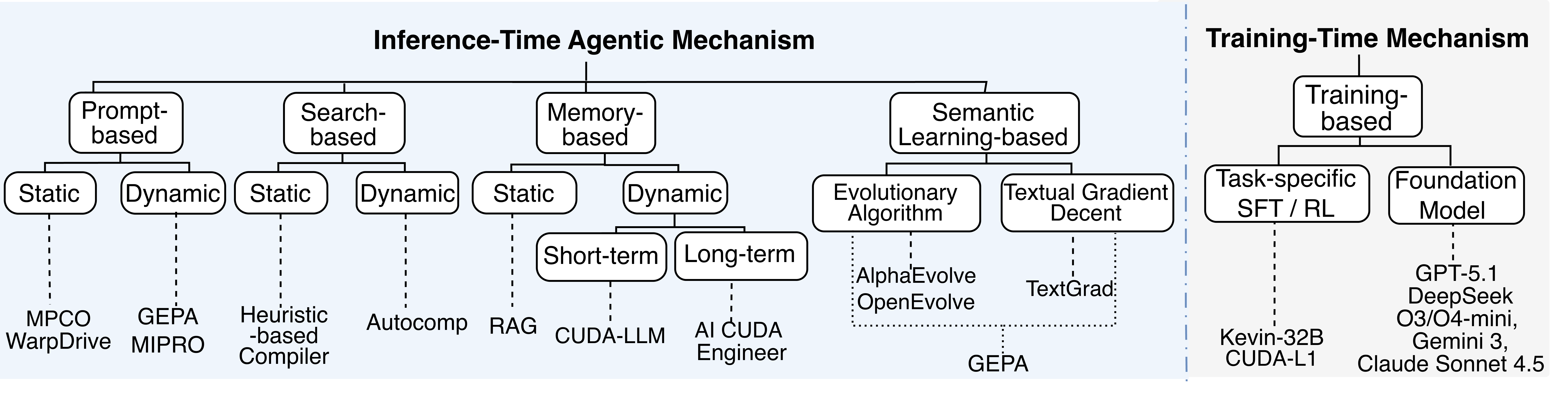}
\caption{Taxonomy of Agentic Flows for LLM-Driven Code Optimization. }
\label{fig:taxonmy}
\end{figure*}

Existing agentic systems for CUDA optimization can be broadly categorized into training-based methods, static prompt engineering approaches, search-based frameworks,  memory-augmented systems, and in-context reinforcement learning techniques. While these approaches have demonstrated the potential of LLM-driven kernel optimization, they vary significantly in adaptability, sample efficiency, cost range, and cross-task generalization.

\paragraph{\textbf{Training-Based Solutions:}} 
The mainstream solution of specializing LLMs for CUDA optimization is via retraining or fine-tuning the models themselves. Examples include the Kevin32B Multi-Turn RL solution for kernel generation~\cite{cognition2025kevin32b}, which uses an iterative feedback loop to generate kernels, calculates a reward, and then uses learned experience and feedback to train the model. Works such as CUDA-L1 extend upon RL training methods by also storing and retrieving a record of past solutions \cite{li2025cuda}. While these are promising approaches that learn from experience and do not require manual prompt engineering, training a model is often an expensive task and potentially impossible in the case of closed-weight models.

\paragraph{\textbf{Static Prompt Engineering Solutions:}}
Early agentic systems for CUDA optimization rely on fixed prompts augmented with manually engineered heuristics. These approaches require substantial effort and domain expertise to construct effective prompts~\cite{opsahl-ong2024optimizing,Damani2024Warpdrive}. While they demonstrate that LLMs can assist in code optimization, their primary limitation is the inability to learn from experience. The prompt remains unchanged regardless of prior outcomes, often leading agents to repeat similar transformations and converge to locally optimal but globally suboptimal solutions~\cite{Ouyang2025fastkernels}. Moreover, adapting to new GPU architectures or specialized domains frequently necessitates additional manual prompt design, particularly in low-resource settings where optimization heuristics are difficult to generalize.

Several systems industrialize this idea via explicit multi-agent workflows. CudaForge, for instance, pairs a Coder and a Judge agent and uses hardware feedback (e.g., profiler metrics) to guide iterative refinement~\cite{zhang2025cudaforge}. However, the overall optimization behavior is still largely dictated by an engineered workflow specification rather than an experience-driven policy that improves across tasks, which constrains adaptability as kernels, domains, and architectures shift. Finally, systems such as KernelFalcon operate over higher-level kernel languages (e.g., Triton) to preserve framework semantics while generating optimized kernels~\cite{wang2025kernelfalcon}. Although these methods benefit from higher-level abstractions and toolchains, but that abstraction can limit the degree of low-level control and fine-grained hardware visibility in low-level CUDA optimization. Besides, they depend on compiler infrastructures, which limit the adaptivity to other systems.   

\paragraph{\textbf{Search-Based Solutions:}} 

More recent systems move beyond static prompting toward dynamic, search-driven optimization. Search-based systems generate candidate kernel variants and leverage iterative sampling and evaluation to select high-performing variants within a single optimization task ~\cite{chen2025cuda_llm}. Beam search techniques improve the search efficiency relative to exhaustive enumeration by pruning weak trajectories ~\cite{hong2025autocompllmdrivencodeoptimization}. These systems enhance agents' exploration efficiency by performing a broader search than relying on a fixed heuristic. 

Despite these advances, the primary limitation of search-based methods is that optimization remains largely kernel-local: each new kernel triggers a fresh, resource-intensive search, and the system typically does not retain or reuse lessons learned from prior kernels ~\cite{chen2025cuda_llm}. In practice, many search procedures require extensive sampling to discover high-performing variants~\cite{hong2025autocompllmdrivencodeoptimization}.  Because they do not track long-term memory (each kernel optimization is treated independently), the agent must rediscover effective strategies for every new task from a default initial state \cite{chen2025cuda_llm}. This results in excessive samples required to rediscover high-performing strategies, and limits optimization decisions to those originally enumerated in the search space.

\paragraph{\textbf{Iterative Refinement on Prompting Policy:}} 
Prompt-policy refinement treats prompts as mutable parameters and improves them using downstream metrics without weight updates. MIPRO optimizes free-form instructions and few-shot demonstrations across multi-stage language-model programs under a black-box objective (no module-level labels or gradients), and reports accuracy gains up to 13\% over baselines on several programs~\cite{opsahl-ong2024optimizing}. GEPA extends this direction with reflective prompt evolution and Pareto-based selection: it analyzes rollout traces, proposes and tests prompt mutations, and achieves large quality gains with far fewer rollouts than weight-space RL, including reported benefits as an inference-time search strategy for code optimization~\cite{agrawal2025gepa}. These optimizers reduce manual prompt engineering, but they typically require re-optimization when the task distribution, tools, or constraints (e.g., new hardware targets) change.

Compared to training-based and search-based solutions, which require significant samples for updating weights, directly updating prompts is significantly more sample-efficient. However, the benefits of GEPA mainly apply to a single task being optimized; although such systems can refine a particular prompt, this process needs to be repeated from scratch for new tasks, limiting generalization from a single training run. In particular, this limitation is due to storing knowledge directly in a series of prompt candidates. This prevents additional knowledge that can be used for further prompt refinement, which is not directly usable during task execution.

\paragraph{\textbf{Memory-Augmented Solutions:}} 
A more sophisticated line of work incorporates long-term memory to learn from experience across multiple tasks to guide future decisions. The AI CUDA Engineer implemented this idea in a staged pipeline to archive correctness-verified kernels and performance data to seed embedding-based retrieval~\cite{Lange2025AICUDAEngineer}. KernelEvolve maintains a hierarchical knowledge base of past kernels with runtime-conditioned search~\cite{liao2025kernelevolve}.

While these methods represent a significant step forward, their effectiveness depends on what is retained and how retrieval is targeted. Most work leverages explicit knowledge distillation. For example, AI CUDA Engineer’s released archive focuses on verified kernels and uses embedding-based retrieval to select in-context exemplars, while KernelEvolve explicitly uses runtime-conditioned retrieval and a structured knowledge-base hierarchy to avoid irrelevant context and preserve context-window efficiency~\cite{Lange2025AICUDAEngineer,liao2025kernelevolve}. More generally, many memory-augmented approaches still rely on the LLM to infer transferable optimization principles from retrieved artifacts at inference time, leaving open the question of how to represent and retrieve bottleneck-level knowledge that transfers across kernels whose surface structure differs.

\paragraph{\textbf{Evolutionary Algorithms.}}
Evolutionary coding agents maintain a population of candidate programs and iteratively improve them using LLM-driven variation operators and evaluator feedback (tests, metrics, or verifiers). Compared to memory-based solutions that explicitly store learnings in a separate data structure, evolutionary approaches implicitly store a past history by curating a population of candidate solutions. AlphaEvolve implements this pattern with an explicit evolutionary loop: sampled ``parent'' programs and inspirations are drawn from a program database to build rich prompts; proposed code differences are evaluated by task-specific evaluators; and promising solutions are added back into the database to drive iterative improvement~\cite{novikov2025alphaevolve}. CodeEvolve provides an open-source realization of this design space, coupling an islands-based genetic algorithm with modular LLM orchestration and selection guided by execution feedback and task-specific metrics~\cite{assumpcao2025codeevolve}. OpenEvolve extends MAP-Elites-style quality-diversity evolution with an explicit program database and an artifact side-channel that propagates execution errors and diagnostics into later generations, reducing repeated failure modes during exploration~\cite{openevolve2025}. 

Storing full code artifacts (e.g., OpenEvolve-style program archives) can incur nontrivial storage overhead and inflate the amount of context needed to condition generation ~\cite{openevolve2025, Lange2025AICUDAEngineer}. Moreover, because these systems often emphasize retaining and sampling high-performing elites, negative outcomes (e.g., slowdowns) may be less systematically represented than they could be for learning robust optimization principles. Finally, retrieval policies are commonly driven by stored metadata or similarity signals, which can miss opportunities to transfer bottleneck-level insights across kernels whose surface forms differ.

While these methods support broad exploration and strategy diversity, their cost scales with evaluator throughput and the robustness of correctness and performance measurement under noisy runtime conditions. Another recent work, EvoEngineer, partitions its memory into solution techniques and population management to handle performant and under-performing approaches \cite{guo2025evoengineer}. However, this approach does not manage long-term memory; instead managing per-problem data structures, which do not exploit cross-problem learning.

\paragraph{\textbf{Semantic Learning via Textual Gradient Decent}}
A significant milestone in enabling learning in agentic systems is to model updates to prompts as approximations of gradients in a textual/semantic space. Instead of applying gradients during backpropagation to model weights, semantic feedback can be propagated back to system prompts using natural language, approximating numerical gradients. This concept, introduced by TextGrad \cite{yao2024textgrad}, enables significantly faster traversal over the parameter space, as natural language feedback contains far denser feedback signals compared to numerical updates. However, TextGrad has not been applied to code optimization domains, which require nuanced hardware-aware feedback. Furthermore, although TextGrad provides a method for textual gradient approximation, it leaves the methodologies of building agentic learning systems that utilize this methodology as an open problem \cite{yao2024textgrad}.

\paragraph{\textbf{In-Context Reinforcement Learning (ICRL):}}

As agentic systems evolved beyond static prompts and search-based methodologies, solutions diverged into memory-augmented methods and learning-based methods. However, both methods are limited by their drawbacks: memory-augmented methods reach scalability bottlenecks, and prior learning-based approaches specialize in particular problems rather than cross-task learning. This paper introduces the emerging field of ICRL to CUDA code optimization. ICRL addresses these bottlenecks by enabling reinforcement learning \emph{at inference time}: the agent conditions on past interactions (i.e., a latent space of states, actions, and rewards) and adapts its behavior \emph{without changing any model parameters}. All adaptation happens in the forward pass by reasoning over the provided history, treating the recent transcript as working memory to infer the task and perform exploration, credit assignment, and policy improvement purely through context processing~\cite{monea2024llmicrl}. Mechanically, the context is a structured log of $(s_t, a_t, r_t)$ tuples, goals, and episode rollouts. The agent parses this history to (i) infer task structure and reward-relevant features, (ii) balance exploration versus exploitation, and (iii) attribute credit by connecting delayed rewards to earlier choices. Lightweight tools such as tables, advantage summaries, natural-language rationales, and rolling statistics can make this explicit, synergizing with policy iteration in context ~\cite{brooks2022policy,demircan2024td,wang2024tdtransformers}.

%% file: mlsys2025style/03_methodology.tex
\section{Methodology}

To address the limitations mentioned above in section~\ref{sec:related}, \sys leverages ICRL, combining the benefits of dynamic search and learning from experience without requiring model training. {\sys} learns from both successes and failures across different optimization problems and can discover both individual optimizations and implicitly encode probabilistic sequences of optimizations via sequences of state transitions. Additionally, we propose a compact, domain-specific long-term memory data structure to store, update, and retrieve distilled knowledge accumulated from previous attempts. {\sys'} ability to consolidate learnings across multiple optimization attempts enables more efficient GPU resource utilization and faster code execution compared to conventional GPU code optimization systems.

\begin{figure}[h!]
\centering
\includegraphics[width=\linewidth]{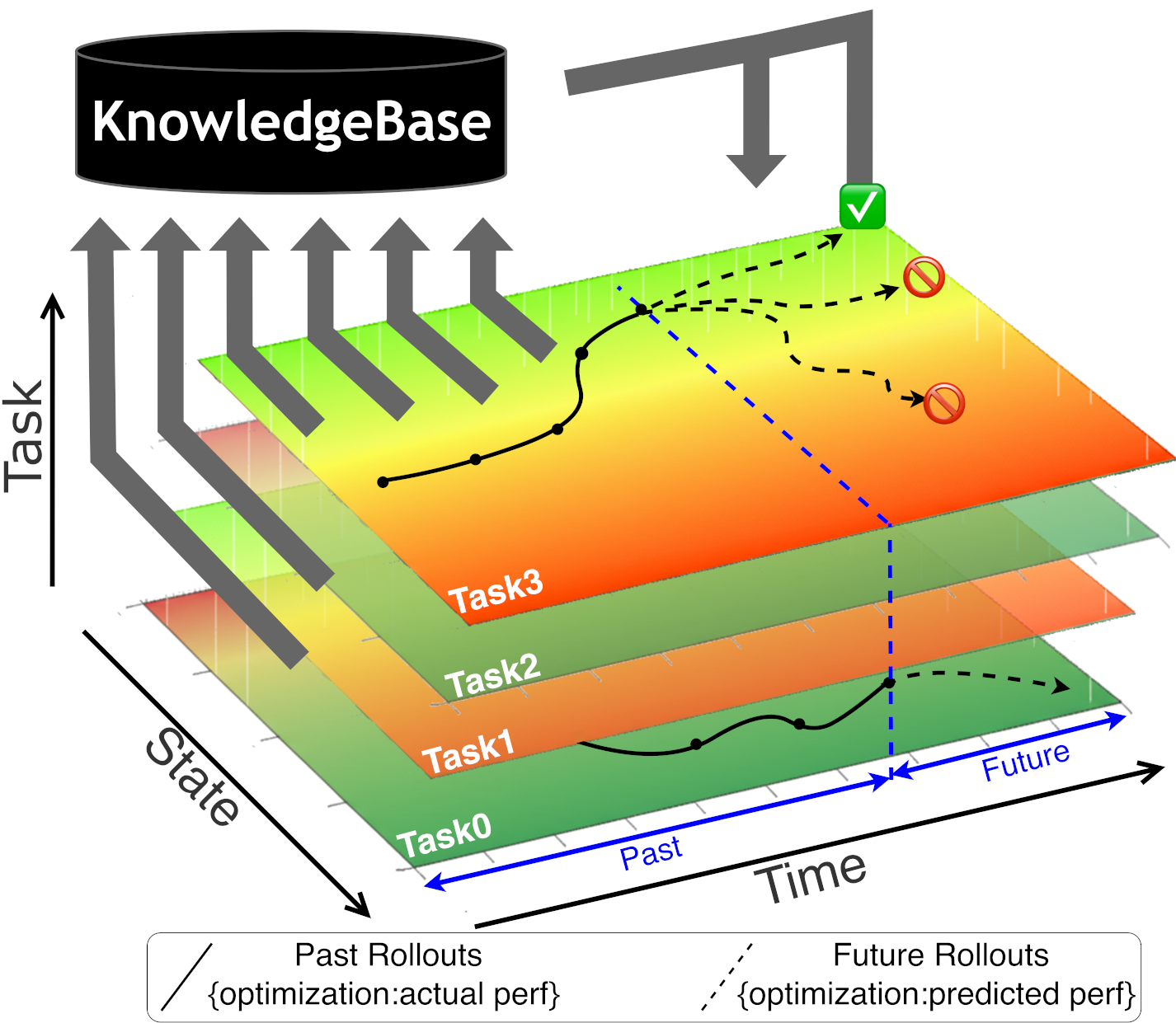}
\vspace{3pt}
\caption{Conceptual Model of Memory-Augmented In-context Reinforcement Learning (MAIC-RL) Across Tasks and Time.}
\label{fig:trajectory}
\end{figure}

Figure~\ref{fig:trajectory} provides a conceptual visualization of how \sys accumulates experience and generalizes knowledge across different optimization tasks over time. The framework operates across three dimensions: State, representing the performance signature of a given CUDA kernel (e.g., memory-bound, compute-bound, etc.); Time, representing the progression of optimization attempts; and Task, representing distinct and unrelated optimization problems.

As the agent undertakes a series of tasks (Task 0 through Task N), it generates optimization trajectories, or past rollouts, on the State-Time plane for each task. Each point on these trajectories corresponds to an optimization attempt and its measured actual performance. This empirical data is continuously distilled and integrated into the central \KB, represented by the upward arrows.

The \KB, acting as the agent's long-term memory and policy, aggregates the learned knowledge. When faced with a new, unseen task, the agent queries the \KB(downward arrow) to generate future rollouts. These future attempts are guided by predicted performance values derived from accumulated experience across past tasks. This process illustrates the system's core capability: it does not solve each problem in isolation but learns generalizable optimization principles that allow it to make more informed, efficient decisions on future, unseen problems.

\begin{figure}[h!]
    \centering
    \includegraphics[width=1.0\columnwidth]{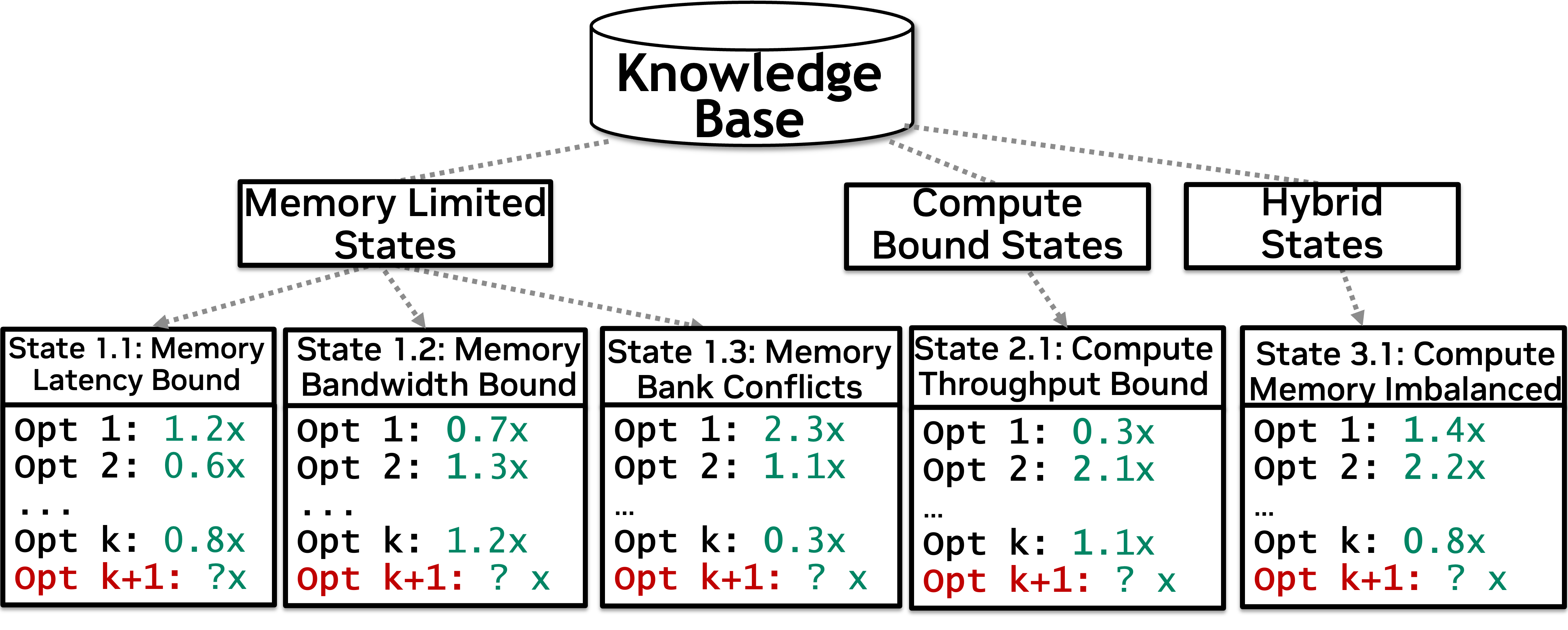}
    \caption{\KB Construction.}
    \label{fig:Knowledgebase}
\end{figure}

\begin{figure*}[!t]
\centering
\includegraphics[width=0.95\textwidth]{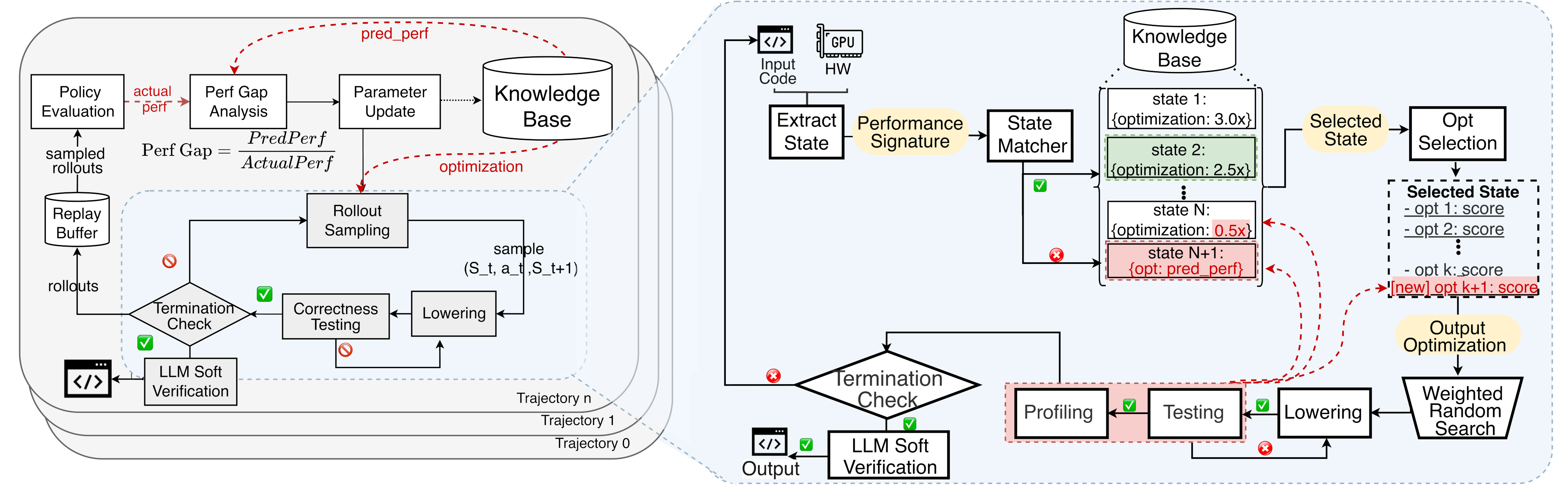}
\caption{Architectural Diagram of \sys. The sub-diagram on the left demonstrates the outer loop of the ICRL  Process: $a_t$ denotes the optimizations generated by the \sys's MAIC-RL policy, $s_t$ denotes unoptimized code prior to $a_t$, and $s_{t+1}$ represents the optimized code after $a_t$ is applied to state $s_t$.  The sub-diagram on the right demonstrates the inner loop of \sys, which runs a single optimization rollout.} 
\label{fig:block_diagram}
\end{figure*}

Our agentic workflow, shown in Figure~\ref{fig:block_diagram}, consists of the following components. 
\begin{itemize}
\item A \textit{Persistent CUDA Knowledge Base} that stores entries of the form 
$\langle \text{state}, \langle \text{optimization}, \text{score} \rangle \rangle$.

\item An LLM-powered \textit{State Extractor} that derives a performance signature from runtime profiling information.

\item A \textit{State Selector} that retrieves relevant $(\text{state}, \text{optimization})$ pairs from the \KB.

\item An \textit{Optimization Selector} that identifies candidate transformations and performs a weighted search to select the top-$k$ optimizations for sampling.

\item A \textit{Lowering Agent} that implements and validates the selected optimization.

\item A \textit{Policy Evaluation} module that analyzes the performance of the generated code and updates the \KB accordingly.
\end{itemize}

\begin{figure}
    \centering
    \includegraphics[width=1.0\columnwidth]{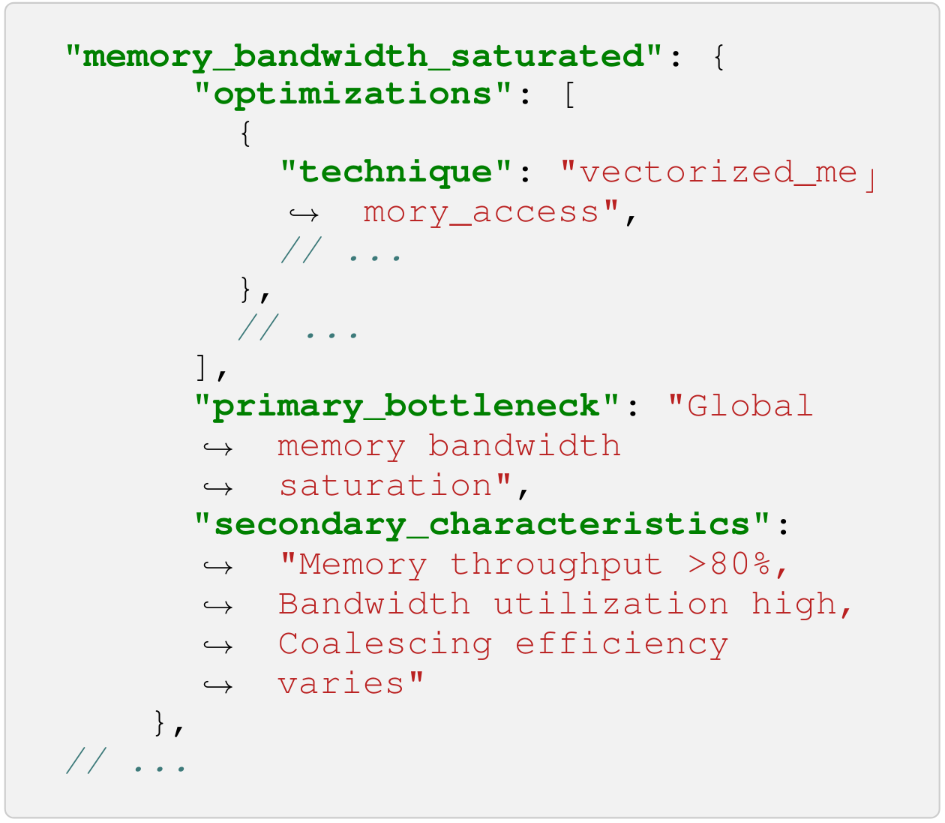}
    \caption{A sample of the \KB data structure, specifying discovered states.}
    \label{fig:kb-format}
\end{figure}

For each new task, we first extract program state from the input program code based on performance characteristics. We provide an example of states within the \KB in Figure~\ref{fig:kb-format}.
 Next, the state-matcher analyzes the performance and code signatures of a kernel and classifies it as a known or discovered state; this comparison uses the performance information for every executed kernels from the ``Details'' section of an Nsight Compute (NCU) report, and compares it against the previously documented primary and secondary bottlenecks of the selected performance state. If it is a discovered state, the agent appends the new state to the \KB.  Otherwise, it uses the state as the key to retrieve a set of candidate optimizations from the \KB. If no optimizations exist yet, it proposes and adds a new set of candidate optimizations to the state in the \KB. From the set of candidate optimizations, the optimization selection agent performs a random weighted selection based on predicted performance gain from the \KB to select the top-k optimizations. The random search ensures that the agent does not always select the best past performer and explores new optimizations. Next, we iteratively explore each of the chosen optimizations by applying them to the initial program and testing and profiling to ensure correctness and performance. Furthermore, this process implicitly discovers successful sequences of optimizations, as different optimization candidates are discovered as the program instances traverse performance states. Finally, we update the \KB with the performance feedback of the optimized kernel.

Our ICRL algorithm is depicted in Algorithm~\ref{alg:ICRL}. Our approach adapts Algorithm~\ref{alg:PG}. REINFORCE \cite{williams1992simple}, a foundational policy-gradient approach in reinforcement learning, to in-context learning. Fundamentally, instead of applying a policy gradient to a set of model parameters, we instead treat the prompt to an LLM-agent as our mutable model parameters, $\theta$. In this case, $\theta$ is a \KB consisting of a set of performance optimization strategies coupled with their expected performance gains. We propose a novel approach in which, instead of directly back-propagating the loss function through the policy to compute the gradient, we use an LLM agent to reason about the policy's performance discrepancy on new samples.

We approximate the policy gradient with two agents, \texttt{PolicyEvaluation} and \texttt{PerfGapAnalysis}, as depicted in Figure~\ref{fig:block_diagram}. After we collect new samples (consisting of tuples of optimized code, optimization actions, and profiling metrics) into a replay buffer, \texttt{PolicyEvaluation} compares the achieved performance of optimizations compared to their expected behavior, and summarizes the key differences in natural language. \texttt{PerfGapAnalysis} then reasons about why the performance results differ and what assumptions were incorrect. Next, we approximate policy update by integrating these changes into the \KB, using another agent, \texttt{ParameterUpdate}. Figure~\ref{fig:block_diagram} shows this process in greater detail.

Essentially, by leveraging LLM agents, we can extract dense performance information in natural language and perform significant analysis during each gradient approximation update. The reward $r_t$ for a sample will include both high-level performance gains, as well as low-level performance breakdowns from GPU profiling tooling. Furthermore, by integrating a reinforcement-learning-based approach, we ensure that subsequent optimizations are informed by real profiling metrics rather than the priors used to generate the initial prompt, $\theta_0$.

\begin{table}[!h]
\caption{Comparison of Classical Policy Gradient and the In-Context {\sys} Approach}
\label{tab:rl_comparison}
\begin{tabularx}{\columnwidth}{@{}lX@{}}
\toprule
\textbf{Component} & \textbf{In-Context {\sys} Analogue} \\
\midrule

\textbf{Policy ($\pi_{\theta}(a_t|s_t)$)} 
& An LLM agent ($\pi$) that, given state $s_t$, generates a distribution of optimized code ($a_t$) based on its context ($\theta$). \\
\addlinespace

\textbf{State ($s_t$)} 
& The unoptimized code prior to applying an optimization. \\
\addlinespace

\textbf{Action ($a_t$)} 
& The optimizations generated by LLM. \\
\addlinespace

\textbf{Next State ($s_{t+1}$)} 
& The optimized code after action $a_t$ is applied to state $s_t$. \\
\addlinespace

\textbf{Reward ($R$)} 
& The reward is a function of the discrepancy between predicted performance and actual performance, where actual performance is measured from running the final generated code on the GPU. \\
\addlinespace

\textbf{Parameters ($\theta$)} 
& The natural language context (the \KB) that guides the LLM. \\
\addlinespace

\textbf{Gradient Calculation} 
& An LLM summarizes the effectiveness of different optimizations from a replay buffer by calculating the discrepancy, $g_k$. Another LLM agent then generates the performance gap analysis, $p_k$. \\
\addlinespace

\textbf{Gradient Update} 
& An LLM rewrites the context document ($\theta$) based on the summary to favor better strategies. \\

\bottomrule
\end{tabularx}
\end{table}

\begin{figure*}[t]
\centering

\begin{minipage}[t]{0.44\textwidth}
\footnotesize
\captionof{algorithm}{\textbf{REINFORCE}}
\label{alg:PG}
\begin{algorithmic}[1]
\State \textbf{Input:} Initial policy params $\theta_0$, env.~$\mathcal{E}$, policy $\pi_\theta$
\State Initialize replay buffer $\mathcal{D} \gets \emptyset$
\For{iteration $k = 0, 1, 2, \ldots$}
    \State Sample initial state $s_0 \sim \mathcal{E}$
    \State Initialize trajectory $\tau \gets []$
    \State $s \gets s_0$
    \For{step $t = 0$ to $T$}
        \State \highlightblue{Sample action $a_t \sim \pi_{\theta_k}(a_t \mid s_t)$}
        \State \highlightblue{Append $(s_t, a_t)$ to trajectory $\tau$}
        \State \highlightblue{Set $s_{t+1} \gets \texttt{EnvStep}(s_t, a_t)$}
        \State \highlightblue{Set $r_t = R(s_t, a_t)$}
    \EndFor
    \State Evaluate return $R(\tau) = \sum_t r_t$
    \State \highlightyellow{Store $(\tau, R(\tau))$ in buffer $\mathcal{D}$}
    \State \highlightyellow{Estimate policy gradient:}
    \State \highlightyellow{$\nabla_\theta J(\theta_k) \gets \frac{1}{|\mathcal{D}|} \sum_{(\tau, R)} \sum_t \nabla_\theta \log \pi_\theta(a_t \mid s_t) R(\tau)$}
    \State \highlightyellow{Update policy: $\theta_{k+1} \gets \theta_k + \alpha \nabla_\theta J(\theta_k)$}
\EndFor
\end{algorithmic}
\end{minipage}
\hfill
\begin{minipage}[t]{0.54\textwidth}
\footnotesize
\captionof{algorithm}{\textbf{LLM-Based Policy Opt.~via Strategy-Guided Rollouts}}
\label{alg:ICRL}
\begin{algorithmic}[1]
\State \textbf{Input:} Initial \KB$\theta_0$, env.~$\mathcal{E}$, LLM policy $\pi_\theta$
\State Initialize replay buffer $\mathcal{D} \gets \emptyset$
\For{iteration $k = 0, 1, 2, \ldots$}
    \State Sample code task $s_0 \sim \mathcal{E}$ \Comment{Initial code}
    \State Initialize rollout trajectory $\tau \gets []$
    \State $s \gets s_0$
    \For{step $t = 0$ to $T$}
        \State \highlightblue{Generate optimized code $a_t \sim \pi_{\theta_k}(a_t \mid s_t)$}
        \State \highlightblue{Append $(s_t, a_t)$ to trajectory $\tau$}
        \State \highlightblue{Set $s_{t+1} \gets a_t$ \Comment{Set new code as state}}
        \State \highlightblue{Set $r_t = R(s_{t+1})$ \Comment{Profile-based reward}}
    \EndFor
    \State Evaluate total reward $R(\tau)$
    \State \highlightyellow{Store $(\tau, R(\tau))$ in buffer $\mathcal{D}$}
    \State \highlightyellow{$g_k \gets \texttt{PolicyEvaluation}(\mathcal{D}, \theta_k)$}
    \State \highlightyellow{$p_k \gets \texttt{PerfGapAnalysis}(g_k)$}
    \State \highlightyellow{$\theta_{k+1} \gets \texttt{ParameterUpdate}(\theta_k, p_k)$}
\EndFor
\end{algorithmic}
\end{minipage}
\vspace{4pt}
\hrule

\end{figure*}

%% file: mlsys2025style/05_evaluation.tex
\section{Evaluation}

\newcolumntype{P}[1]{>{\raggedright\arraybackslash}p{#1}}
\begin{table*}[h!]
\centering
\caption{Comparison among {Ours ({\sys})}, AI CUDA Engineer, and IREE ML compiler baseline.}
\scriptsize
\renewcommand{\arraystretch}{1.35}

\begin{tabular}{@{}P{1.7cm}P{2cm}P{2cm}P{1.1cm}P{3.4cm}P{5.1cm}@{}}
\toprule
{System} &
{Agent's Models} &
{Hardware Setup} &
{Dataset} &
{Hyperparameter} &
{Test Harness} \\
\midrule
\rowcolor{RowBlue}
{Ours} &
GPT-4.1 and GPT-5.0. &
NVIDIA A6000, A100, H100, L40S GPUs. &
KernelBench Level 1-2, Subset of Level 3 & 
10 iterations, 10 rollout steps per iteration  &
\cellcolor{CellYellow}  \textbf{Functionality Check}: Passing kernels are compiled and executed on GPUs; their outputs are compared to PyTorch baselines with multiple randomized seeds to ensure correctness and prevent overfitting. \\

\rowcolor{RowGreen}
{AI CUDA Engineer} &
O4-mini, Claude 3.7 Sonnet, Gemini 2.5 Pro, GPT-4.1, and O3. &
NVIDIA H100 GPUs. &
\cellcolor{CellYellow} KernelBench Level 1–2 & 
10 generations; 8 proposals sampled per generation; top 4 evaluated. &
\cellcolor{CellYellow}\textbf{LLM Soft-Verification}: The LLM scans each kernel for likely compilation, memory, or numerical errors, filtering out bad code before GPU execution.
\\

\rowcolor{RowGray}
{ML Compiler (IREE)} &
N/A. &
NVIDIA A100 and A6000 GPUs. &
\cellcolor{CellYellow} & 
\texttt{-O3} optimization level with LLVMGPU passes 
&
Kernels are profiled via NCU by wrapping the \texttt{iree-run-module} command, executing VMFB modules with randomized inputs to capture execution traces and performance metrics. \\
\bottomrule
\end{tabular}
\label{tab:cuda_vs_tachyon_iree}
\end{table*}

\begin{table}[h!]
\centering
\caption{Performance Comparison Across GPUs and Datasets}
\label{tab:summary_all}
\scriptsize
\renewcommand{\arraystretch}{1.2}

\begin{tabular}{>{\raggedright\arraybackslash}p{0.7cm}
                >{\centering\arraybackslash}p{0.7cm}
                >{\centering\arraybackslash}p{0.55cm}
                >{\centering\arraybackslash}p{0.7cm}
                >{\centering\arraybackslash}p{0.4cm}
                >{\centering\arraybackslash}p{0.4cm}
                >{\centering\arraybackslash}p{0.4cm}
                >{\centering\arraybackslash}p{0.5cm}
                >{\centering\arraybackslash}p{0.5cm}}
\toprule
\textbf{} & \textbf{ValidRate} & \textbf{Average} & \textbf{GeoMean} & \textbf{Med.} &
\textbf{Min} & \textbf{Max} & \textbf{\%$>$1x} & \textbf{\%$<$1x} \\
\midrule

\multicolumn{1}{l}{\textbf{L40S}} & \multicolumn{8}{c}{\textbf{Level 1}} \\
\midrule
IREE     & 85\% & 0.856 & 0.268 & 0.518 & 0.0035 & 9.01  & 21.18\% & 78.82\% \\
CUDAEng  & 82\% & 1.728 & 1.101 & 1.202 & 0.0659 & 32.35 & 74.39\% & 25.61\% \\
Ours     & 93\% & 4.502 & 1.080 & 1.037 & 0.0801 & 183.96& 59.72\% & 40.28\% \\
\midrule

\multicolumn{1}{l}{\textbf{}} & \multicolumn{8}{c}{\textbf{Level 2}} \\
\midrule
IREE     & 83\% & 1.137 & 0.279 & 0.298 & 0.0255 & 38.50 & 16.87\% & 83.13\% \\
CUDAEng  & 83\% & 3.865 & 1.695 & 1.574 & 0.328  & 123.39& 79.52\% & 20.48\% \\
Ours     & 95\% & 9.419 & 2.214 & 2.074 & 0.0488 & 362.29& 72.60\% & 27.40\% \\
\midrule

\multicolumn{1}{l}{\textbf{}} & \multicolumn{8}{c}{\textbf{Level 3}} \\
\midrule
Ours     & 67\% & 1.749 & 1.502 & 1.894 & 0.529 & 2.681& 75.0\% & 25.0\% \\
\midrule
\midrule

\multicolumn{1}{l}{\textbf{H100}} & \multicolumn{8}{c}{\textbf{Level 1}} \\
\midrule
CUDAEng  & 82\% & 1.935 & 1.025 & 1.085 & 0.0492 & 54.40 & 64.20\% & 35.80\% \\
Ours     & 86\% & 2.817 & 1.497 & 1.137 & 0.0647 & 32.74 & 70.37\% & 29.63\% \\
\midrule

\multicolumn{1}{l}{\textbf{}} & \multicolumn{8}{c}{\textbf{Level 2}} \\
\midrule
CUDAEng  & 82\% & 1.356 & 1.214 & 1.170 & 0.402  & 3.651 & 61.33\% & 38.67\% \\
Ours     & 81\% & 10.223& 2.592 & 2.291 & 0.111  & 213.65& 84.85\% & 15.15\% \\
\midrule

\multicolumn{1}{l}{\textbf{}} & \multicolumn{8}{c}{\textbf{Level 3}} \\
\midrule
Ours     & 67\% & 1.336 & 1.110 & 1.333 & 0.375 & 2.302 & 75.0\% & 25.0\% \\
\midrule
\end{tabular}

\textit{ValidRate indicates the percentage of valid runs that have valid initial CUDA code that successfully pass both functionality and LLM-based soft verification checks. Baseline (1.0x) is measured as the best performance among \texttt{torch.eager} and \texttt{torch.compile}.}
\end{table}

We evaluate four representative optimization systems: Our {\sys},  the AI CUDA Engineer, Kernelsseum \cite{ouyang2024kernelbench}, and the IREE compiler \cite{iree-software} across diverse GPU architectures and benchmark levels. The overall configuration, including used models, hardware targets, datasets, initialization methods, and evaluation metrics, is summarized in Table \ref{tab:cuda_vs_tachyon_iree}. The comparison of compiler- and agent-driven optimization pipelines is under equivalent execution and profiling conditions. 

\subsection{Evaluation Setup}\label{sec:setup}

We evaluate {\sys} against a suite of representative optimization systems, including compiler and agent-based solutions, across diverse GPU architectures and benchmark levels. To ensure a fair comparison, we provide a table with a comprehensive summary of the evaluation configuration provided in Table~\ref{tab:cuda_vs_tachyon_iree}. For our results and IREE, we use the sum of elapsed cycles of all kernels using the NCU profiler for both optimized kernels and the PyTorch baselines. For AI CUDA Engineer, we use the test harness provided with the released kernels, which uses application-level timing for both optimized kernels and PyTorch.
Our method is compared against several baselines: PyTorch’s default eager execution mode; \texttt{torch.compile}, a JIT compiler integrated into PyTorch \cite{pytorch_torch_compile}; the IREE ML compiler \cite{iree-software}; The primary agentic comparison is against the AI CUDA Engineer, a state-of-the-art solution in CUDA code generation.

\subsection{Evaluation Metrics}\label{sec:metrics}
The target metrics include kernel performance and system performance. The efficacy of each optimization system is assessed using a comprehensive set of metrics that capture both the performance of the generated code and the efficiency of the optimization process. Code performance is the primary focus, quantified by the speedup achieved over established baselines, including the original PyTorch implementation and related work. We defined \textit{Valid Rate} as the percentage of optimization problems that successfully pass both functionality and LLM-based soft verification checks. Baseline (1.0x) is measured as the best performance among PyTorch Eager and \texttt{torch.compile}. To provide a holistic view of performance improvements, the mean speedup across all successful runs is reported with baseline (1.0x) measured as the best performance among PyTorch Eager and \texttt{torch.compile}.

Given that comparison groups may generate the same number of correct kernels at different rates, inducing different system costs. We evaluate the $\mathit{fast}_p$ metric to characterize the distribution of high-impact optimizations. $\mathit{fast}_p$ is defined the percentage of kernels that are at least $r$ times faster than the baseline within $k$ attempts ~\cite{ouyang2024kernelbench}, which defines as the fraction of tasks that both produce correct outputs and achieve a speedup—defined as the ratio of PyTorch wall-clock time to generated kernel time-exceeding a threshold \(p\):
\[
\text{fast}_p = \frac{1}{N} \sum_{i=1}^{N} \mathbf{1}(\text{correct}_i \land \{\text{speedup}_i > p\}),
\]
System performance is also evaluated using cost, measured in the total number of tokens consumed to optimize the kernel. In our evaluation, we compare the tokens consumed by \sys's MAIC-RL agent compared to a baseline minimal agentic iteration loop.

\subsection{Execution Harness}
To evaluate our implemented kernels, we use a C++ based test harness, which includes driver code and a reference Torch implementation of the kernel. The driver also launches \sys's optimized CUDA implementation, which consists of one or more innovations to CUDA kernels. First, if compilation errors occur, the solution is discarded, and the compilation feedback is returned to the code-lowering agent. Second, we run the code without profiling and return the numerical verification. Incorrect solutions are also re-attempted. Finally, we execute the optimized code, using NCU to profile kernels used in the optimized code. NCU results, including elapsed cycles and a profiled report of kernels’ detailed characteristics (e.g., memory- vs. compute-dominated behavior and major stall sources). All instances of kernels are profiled and independently reported to the agent in the order they were executed, to account for cases where the same kernel is used multiple times during program execution. 

\subsection{Validation Harness}
Given the emergent challenge of reward hacking in agentic systems, a strong emphasis is placed on Correctness Verification. This phenomenon, where an AI agent achieves a high reward by exploiting unintended loopholes in the evaluation environment rather than by solving the intended task, poses a significant threat to the validity of automated optimization. As reported by AI CUDA Engineer \cite{Lange2025AICUDAEngineer}, which was found to achieve illusory speedups by exploiting a memory bug in the evaluation code to bypass correctness checks. To guard against such failure modes, generated code is numerically validated against the original PyTorch implementation. Additionally, we employ an LLM-based soft-verification pass that validates the structure of the final kernel against the initial KernelBench PyTorch implementation. Critically, this detects scenarios where the optimization agent eliminates functionality from the original kernel, leading to incorrect speedups. Additionally, to prevent shortcut behaviors such as calling optimized external libraries, our soft-verification agent ensures that generated kernels only use native CUDA functionality.

\subsection{Comparison against PyTorch}

\begin{figure*}[!h]
  \centering
  \includegraphics[width=.65\linewidth]{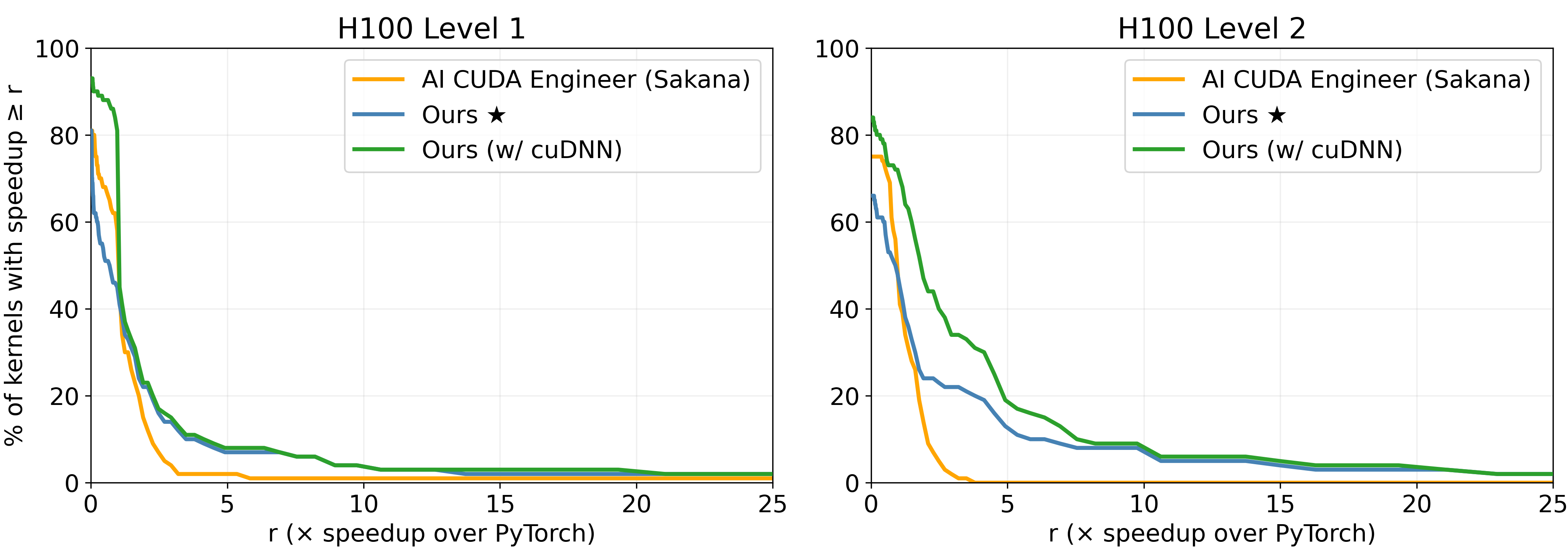}
  \caption{\vspace{2pt} fast\_p($r$) distributions on NVIDIA H100 for KernelBench Level~1 and Level~2. The plot shows the percentage of kernels achieving at least $r\times$ speedup over PyTorch. Our approach yields a larger fraction of kernels with moderate-to-high speedups, particularly on Level~2 workloads that require coordinated, multi-step optimizations rather than isolated local rewrites.}
  \label{fig:h100_fastp}
\end{figure*}

\begin{figure*}
    \centering
    \includegraphics[width=0.7\linewidth]{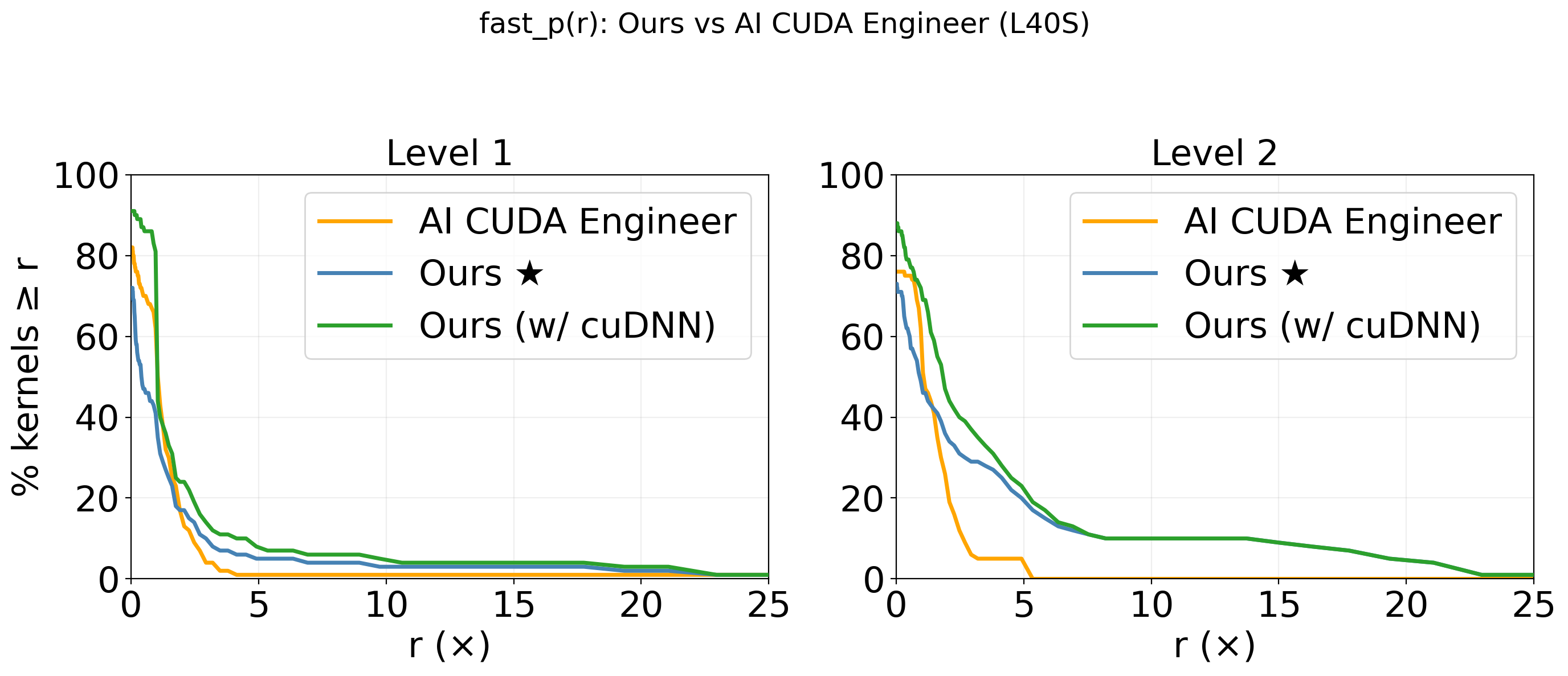}
    \caption{$\mathit{fast}_p$ curves for AI Cuda Engineer and {\sys} on an L40S GPU, including cuDNN augmentations.}
    \label{fig:fastpr_sakana_ours}
\end{figure*}

Our primary baseline is a comparison against a PyTorch baseline, running the original KernelBench baseline code. As described in Section~\ref{sec:metrics}, we use the $\mathrm{fast}_p$ metrics to demonstrate what percentage of kernels improve by the desired performance target over our baseline. On both Level 1 and Level 2 benchmarks, over 50\% of optimized solutions improve upon the best-performing result between native Eager PyTorch and PyTorch Compile. This effect is particularly pronounced in Level-2 benchmarks, as seen in Figure~\ref{fig:fastpr_sakana_ours}. The reason we observe more significant benefits for Level 2 code is that these kernels typically have multiple composed operators, providing a larger search space for optimizations that the agentic flow can exploit. In comparison, Level 1 kernels are much simpler, limiting the scope of optimization approaches.
\begin{figure*}[h!]
    \centering
     \includegraphics[width=0.65\linewidth]{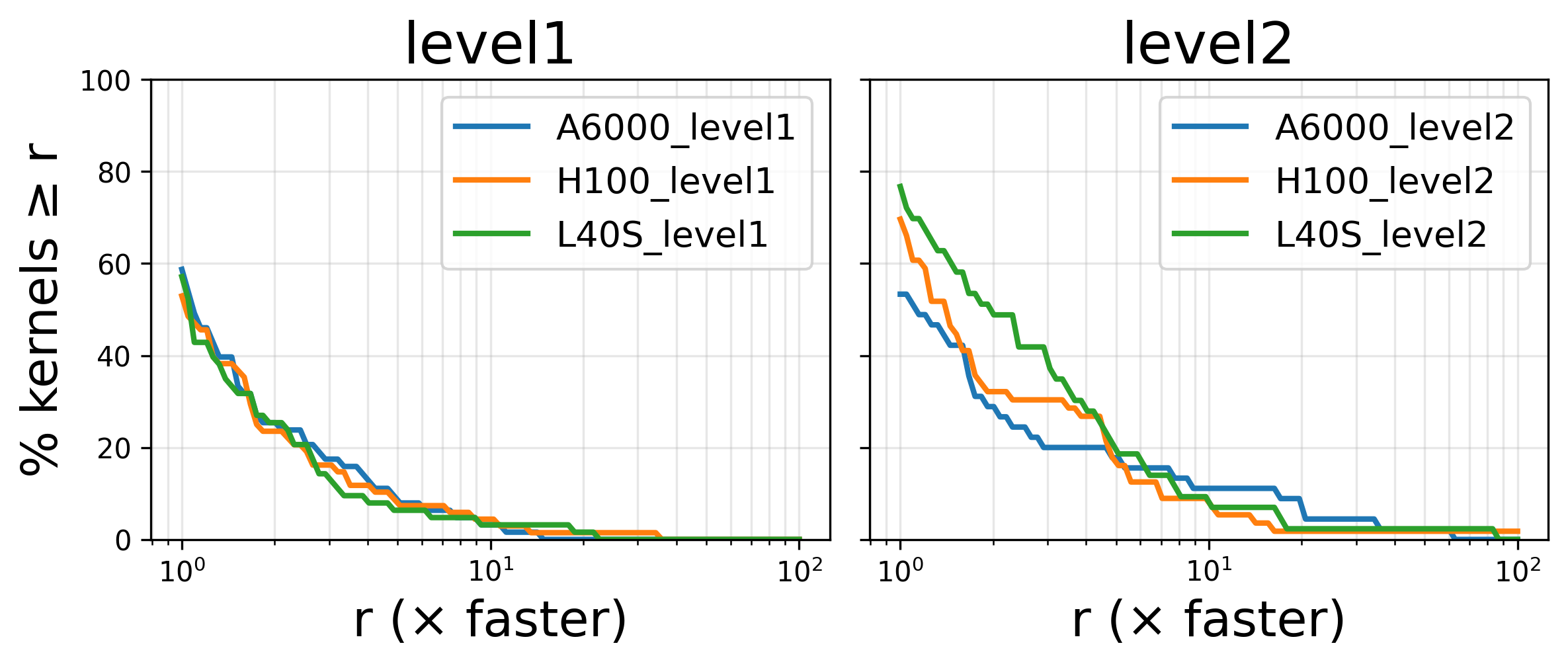}
    \caption{$\mathit{fast}_p$ curve for {\sys}'s performance gains versus initial CUDA code across four NVIDIA GPU architectures: A6000 A100 (Ampere), H100 (Hopper), and L40S (Ada Lovelace)  running KernelBench Level 1 and Level 2 Problems over Naive CUDA.}
    \label{fig:vs-GPU}
\end{figure*}

\subsection{Comparison against Naive CUDA}
Comparison with PyTorch demonstrates \ sys's capability against a standard reference. However, \sys's optimization flow does not directly optimize on top of native PyTorch code, but on a prior CUDA implementation.
{\sys} begins optimization from functionally correct CUDA kernels generated from the KernelBench PyTorch implementations via an LLM agent. After verifying the correctness of the generated kernel and driver, we execute our optimization workflow. We plot $\mathit{fast}_p$ curves across several GPUs against the naive CUDA code in Figure~\ref{fig:vs-GPU}. When comparing against the naive CUDA baseline, we observe significant improvements, up to 100$\times$ over the baseline. However, this is primarily due to the functional baseline missing basic optimization techniques, such as tiling or vectorization. Due to this, we observe the greatest speedup in the simple Level 1 kernels, which can benefit greatly from these techniques.

\subsection{Comparison Against Other Agentic Workflows} Figure~\ref{fig:fastpr_sakana_ours} shows the fastp results for AI Cuda Engineer on the L40S GPU. {\sys} with cuDNN shows a consistently higher percentage of kernels with speedup exceeding r for both Level 1 and Level 2 problems. Additionally, the AI CUDA Engineer sees a lower validation rate (82\%) and a geometric mean speedup of 1.1x on Level 1 and 1.7x on Level 2 compared to the Pytorch baseline. Note that performance results may be impacted by the difference in our evaluation methodology as described in Section~\ref{sec:setup}. We also compare against Kernelsseum, which represents the best zero-shot generation results from a simple prompting-based approach \cite{ouyang2025kernelbench}. Compared to zero-shot baselines, we observe significant speedups due to the ability of {\sys} to learn from experience. However, future work can consider applications of the \textit{Knowledge Base} to improve the generation of performant kernels without the need for online profiling.

\subsection{Comparison against IREE Compiler} 
To establish a baseline representing modern ML compiler capabilities, we evaluated the IREE compiler~\cite{iree-software}, leveraging the \texttt{torch-mlir}~\cite{torch-mlir-software} frontend, on the KernelBench level~1 and level~2 benchmarks. Targeting NVIDIA A100 and A6000 GPUs, we attempted 400 compilations (100 kernels $\times$ 2 levels $\times$ 2 GPUs). IREE successfully compiled 358 kernels (89.5\%), but 42 attempts failed. These failures primarily stemmed from unimplemented \texttt{torch-mlir} lowerings for numerous PyTorch operations (e.g., \texttt{torch.aten.diag}, \texttt{torch.aten.broadcast\_tensors}), highlighting the significant effort required for ML compilers to keep pace with evolving frontend frameworks like PyTorch.  Tables \ref{tab:summary_all} showcase the slowdown of IREE against our baseline (PyTorch Eager). IREE only achieves 27\% of the speedup of Pytorch Eager for level 1 and 28\% for level 2. This discrepancy can be partly attributed to IREE's current primary optimization focus on AMD GPUs and CPUs, whereas the baseline benefits from more mature and specialized optimizations deeply integrated within the PyTorch and native NVIDIA CUDA ecosystem.

\subsection{Extending to Full Models}

We primarily apply \sys at the operator and level-granularity; this can be scaled to accelerate end-to-end models by optimizing individual components. However, we also evaluate our methodology on a subset of KernelBench Level~3 workloads. We therefore run our full optimization pipeline on a subset of KernelBench Level~3 workloads to assess whether the approach generalizes beyond isolated kernels.

Using the KernelBench Level~3 Network \textit{LeNet5} as an example, our generated CUDA implementation achieves a 2.68$\times$ speedup over the PyTorch baseline, demonstrating that the methodology remains effective when multiple layers are composed into a full model rather than isolated kernels.  For larger level3 blocks, such as \textit{SqueezeNetFireModule}, we achieve 1.95$\times$ improvement over PyTorch.

The generated CUDA code in Section~\ref{sec:appendix} shows that our agent applies the \KB discovered at Level~1 and Level~2 to Level~3. In particular, the agent performs cross-layer fusion to reduce kernel launch overhead, improves memory locality by reusing activations across consecutive linear layers, and applies algebraic and structural simplifications that reduce redundant memory traffic. Besides cross-layer fusion, algebraic simplification, and memory locality, these optimizations are learned in Level~1 and Level~2 and transferred to Level~3 kernels. 

However, we observe scaling limitations for optimizing CUDA code for full-model problems. First, due to the large number of kernels and operators within level3 models, processing a singular optimization per iteration limits the performance improvement for the whole model. For problems with multiple diverse kernels with diverse performance states and optimization opportunities, future work should consider processing vectors of state characterization and optimization targets across the model. Furthermore, compared to the relatively concise problem representation of KernelBench Python code, implementing full networks in native CUDA code results in extremely verbose source files, with significant overheads to boilerplate code, limiting reasoning ability, and diluting LLMs' ability to identify performance signals. We expect that the agentic workflow would benefit from pre-processing the problem hierarchically into more manageable sub-problems; given our results in level2 problems, this would improve \sys's ability to improve end-to-end model performance by optimizing fused-layer sub-blocks.

\subsection{Cost Summary}
\begin{figure}
    \centering
    \includegraphics[width=0.8\linewidth]{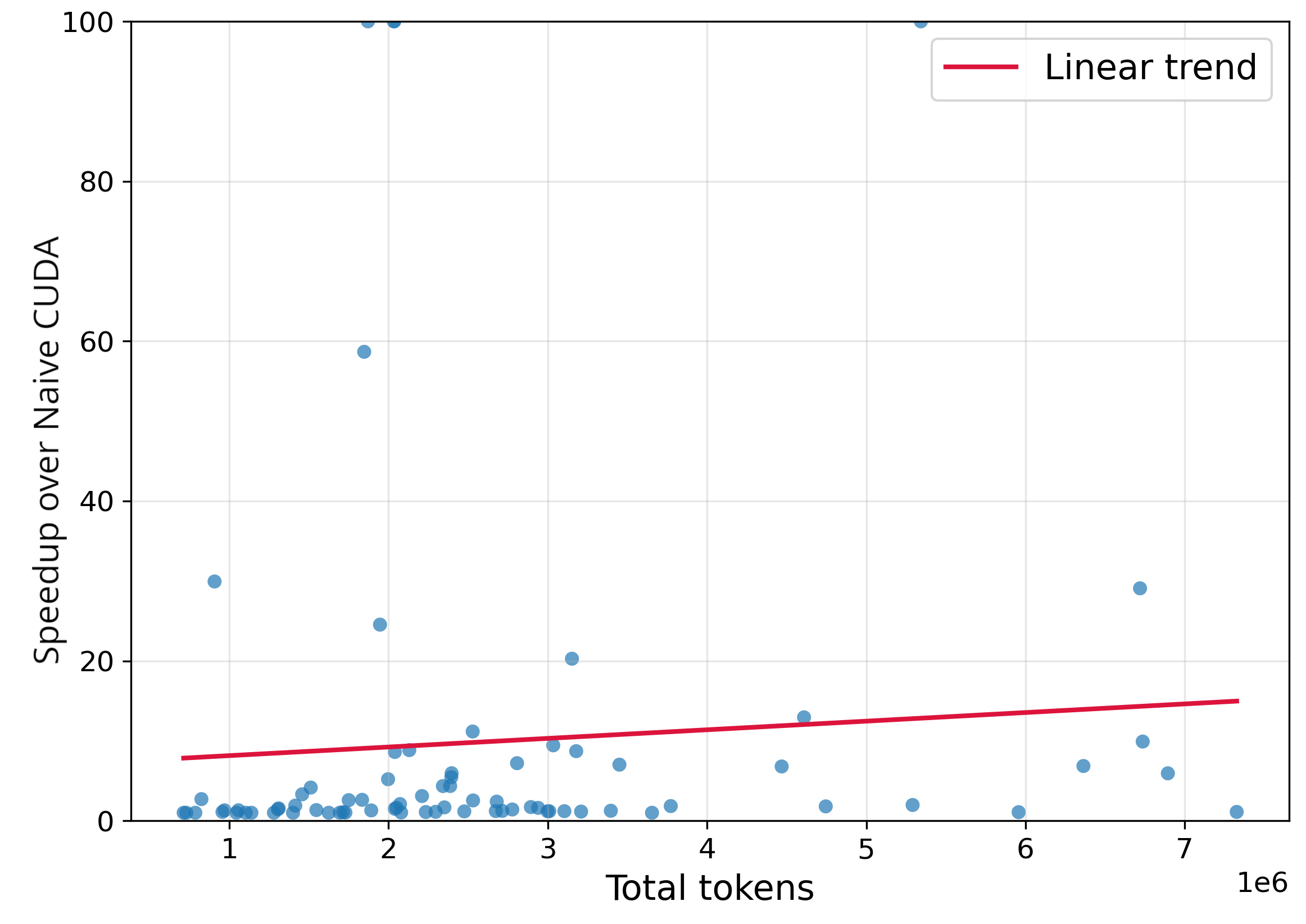}
    \caption{Scatter plot of speedups over the original CUDA per token costs.}
    \label{fig:costs}
\end{figure}

In Figure~\ref{fig:costs} we provide a summary of the measured speedup over the original CUDA reference code achieved per total tokens consumed. Although each problem runs for the same number of iterations, token count can vary widely, in part due to code size, number of kernels profiled, and complexity of optimizations selected. Overall, we observe a positive correlation, gaining better performance gains when more tokens were consumed to process the problem.

\subsection{Performance Summary}

\begin{figure}[t]
  \centering
  \includegraphics[width=\linewidth]{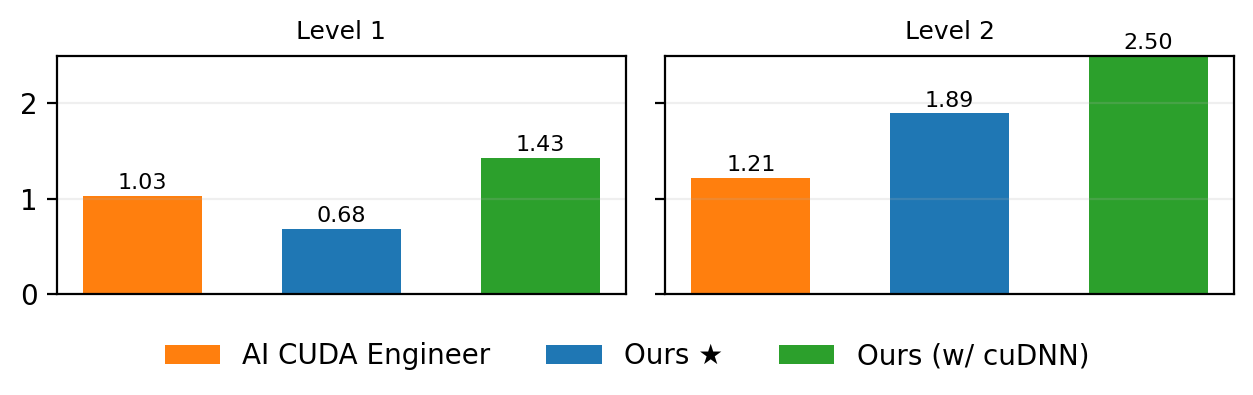}
  \caption{Geometric-mean speedup over PyTorch on NVIDIA H100 for KernelBench Level~1 and Level~2. We compare AI CUDA Engineer, our approach without cuDNN, and our approach with cuDNN enabled. Under identical hardware and evaluation conditions, our method achieves higher average speedups on both levels and composes effectively with vendor libraries.}
  \label{fig:summary-plot}
\end{figure}
Overall, we demonstrate that {\sys} can improve upon a wide range of prior techniques. In Figure~\ref{fig:summary-plot}, we show the speedups over PyTorch across several implementations. First, we observe the greatest speedup of 7.9$\times$ against traditional compiler baselines such as IREE. Furthermore, we have additional speedups over zero-shot solutions due to our ability to learn from experience. For simple kernels with limited available optimization strategies, we achieve similar performance to prior agentic workflows. However, for Level 2 kernels, we are able to achieve improved performance over SOTA agentic workflows due to our ability to apply diverse optimizations. 

%% file: mlsys2025style/06_AB_experiment.tex

\section{Distribution of Optimization Usage}\label{sec:opt-dist}

\begin{figure*}
    \centering
    \includegraphics[width=\linewidth]{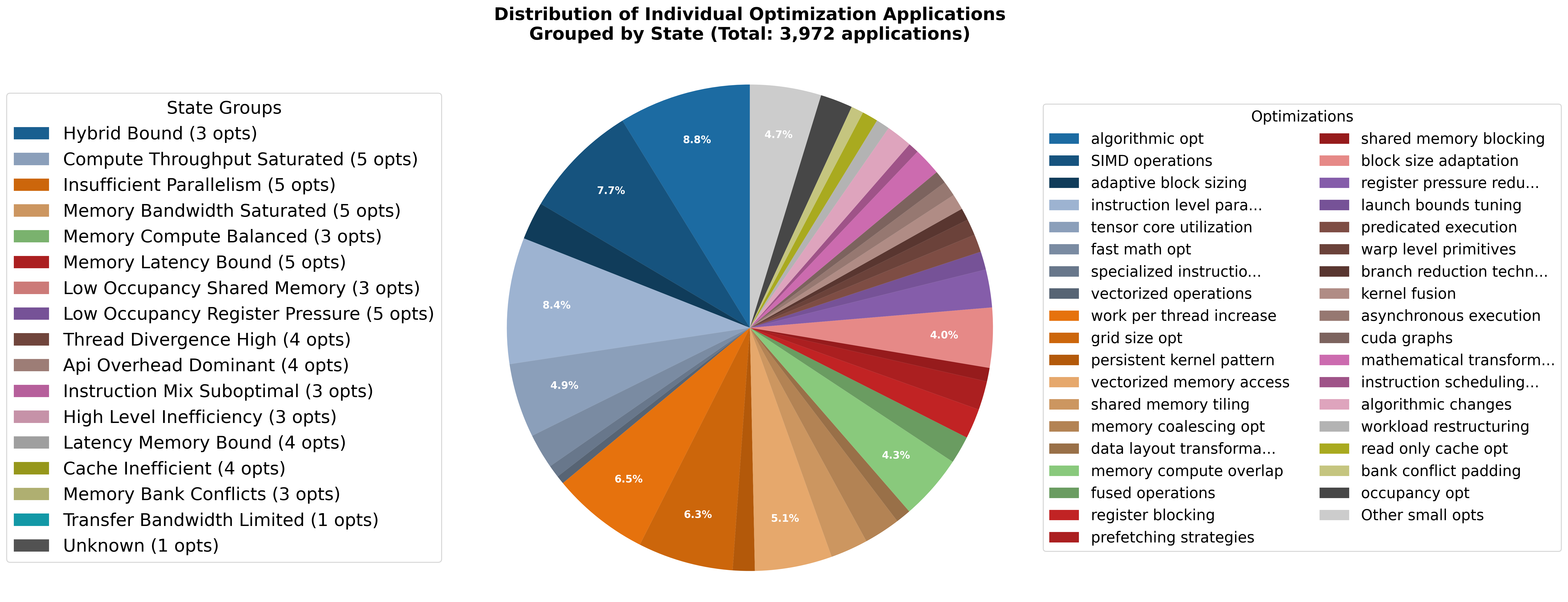}
    \caption{Distribution of 3972 optimization applications by {\sys} to Level 1 and Level 2 KernelBench CUDA kernels on an A6000 GPU.}
    \label{fig:opt_pie}
\end{figure*}

A key benefit of {\sys} is its ability to discover and apply a diverse set of useful optimizations. The learned optimizations are stored in the \KB, approximately 50 KB in size. These optimizations cover a wide range of kernel performance signatures and techniques, as depicted in Figure~\ref{fig:opt_pie}. Optimizations are grouped by state and are aggregated across all successful applications. No state dominates across optimization iterations, with no state exceeding 20\%. An average of 5.5 states is reached per kernel, highlighting the need for detailed profiling feedback, as optimizations can dramatically change the program's kernel signature.

{\sys}'s optimization trajectories provide an opportunity for characterizing directed sequences of optimizations. Analysis of optimization sequences on L40S GPUs reveals a strong, compute-centric strategy dominated by tactics like \texttt{instruction\_level\_parallelism} and \texttt{tensor\_core\_utilization}. This approach features significant repetition, but this ``micro-tuning'' often proves unproductive; for example, over 50\% of repeated \texttt{instruction\_level\_parallelism} applications (and over 80\% for \texttt{grid\_size\_optimization}) yield negligible speedups ($< 1.01\times$). In contrast, the efficacy of transitions between different optimizations reveals valuable ``prep\(\rightarrow\)compute'' patterns. The most significant gains come from preparatory memory optimizations: applying \texttt{shared\_memory\_tiling} before \texttt{tensor\_core\_utilization} yields a substantial median speedup of \(\approx\!2.41\times\). Similarly, transforming data layouts before fusing operations ($\approx\!1.95\times$) or simplifying control flow before tensor core tuning ($\approx\!1.42\times$) provides significant, reliable performance boosts. These insights suggest that {\sys}'s approach to learning <state, optimization> pairs can exploit these high-yield preparatory transitions, steering the search away from low-return repetition and towards these more impactful, structured transformations.

\label{app:in_depth_opt_breakdown}

\begin{figure*}
    \centering
    \includegraphics[width=0.8\linewidth]{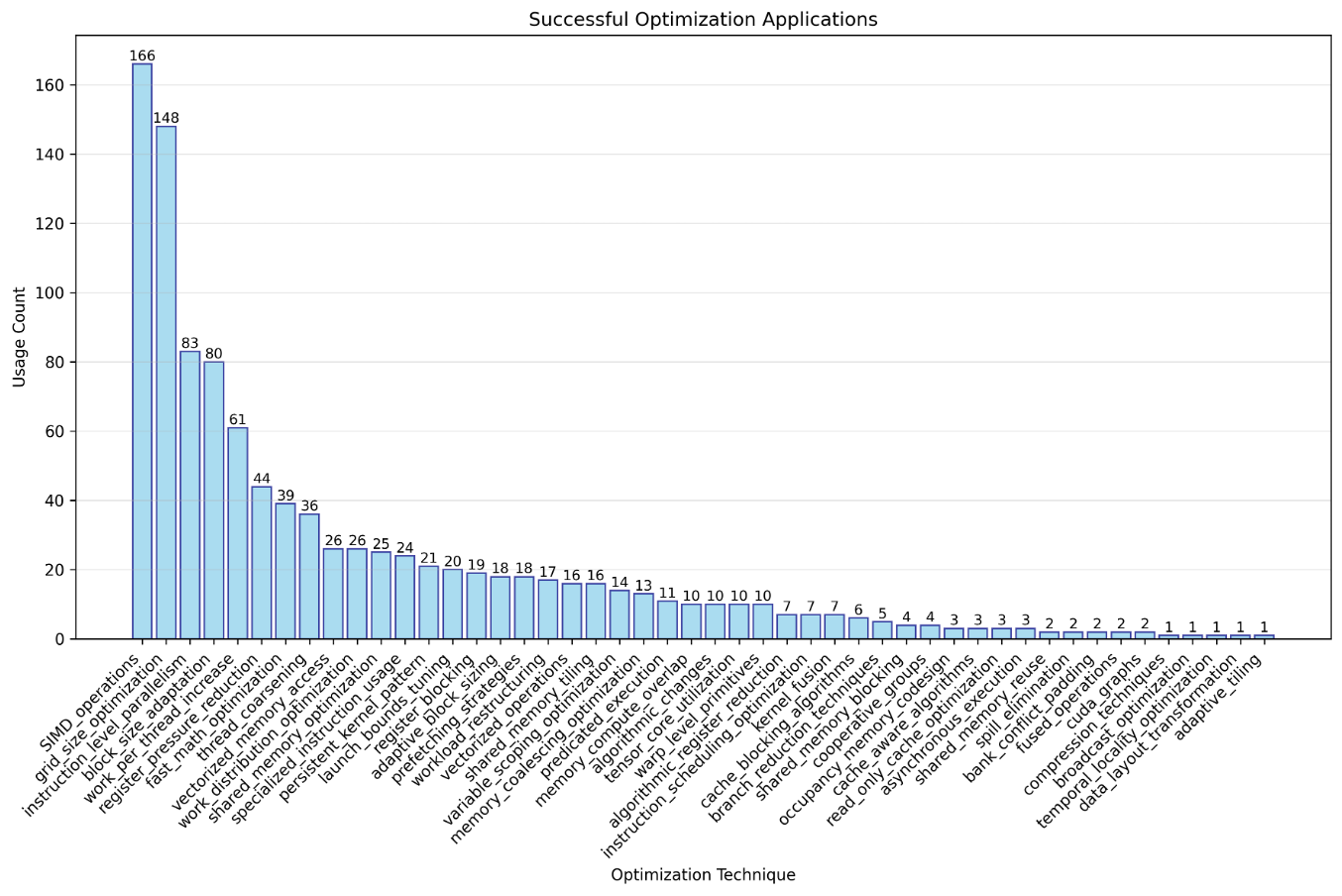}
  \caption{Counts of successful optimization applications per technique.}
  \label{fig:opt_success_counts}
\end{figure*}

\begin{figure*}
    \centering
    \includegraphics[width=0.8\linewidth]{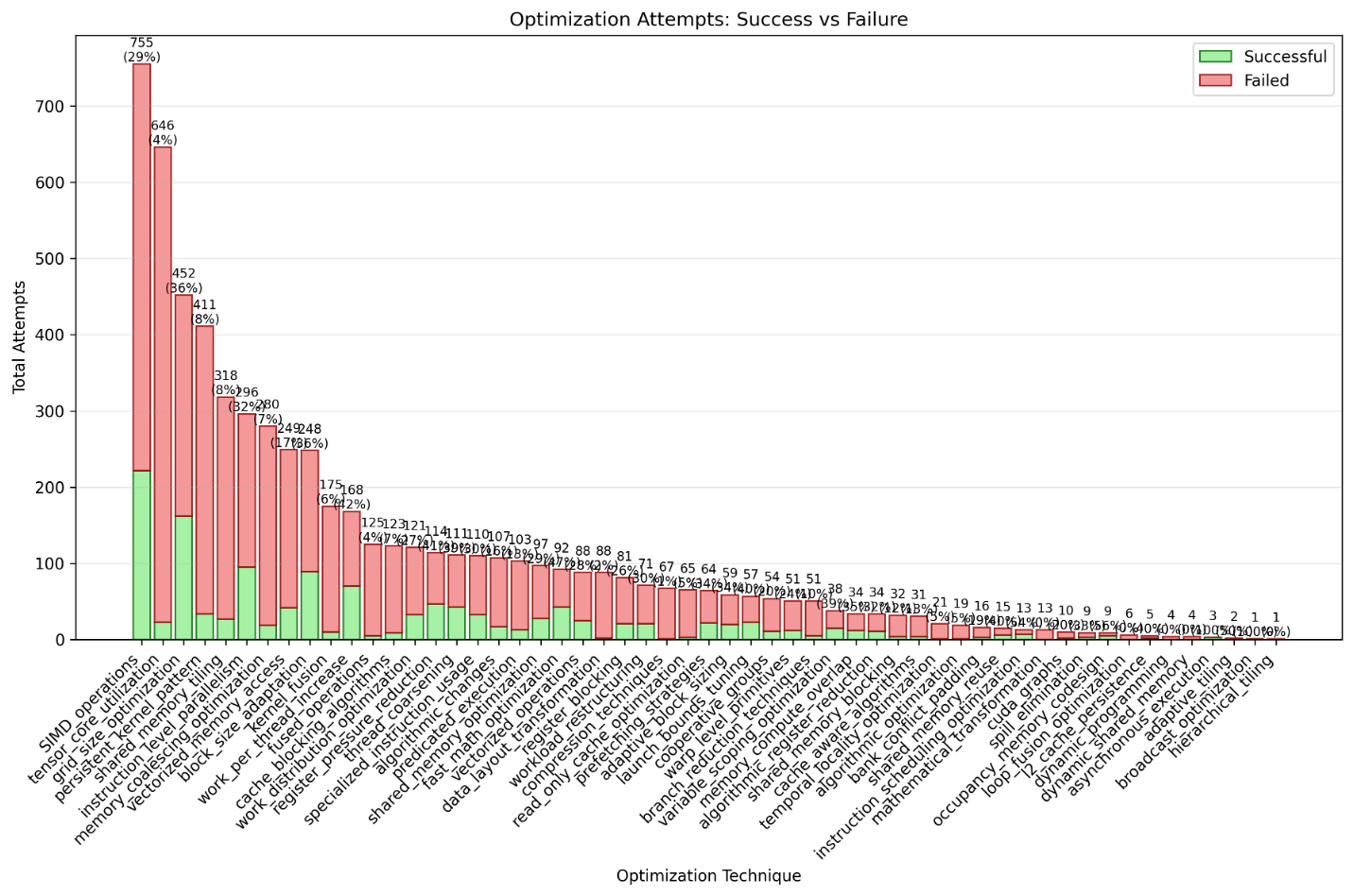}
  \caption{Optimization attempts per technique, stacked by successful versus failed trials.}
  \label{fig:opt_attempts_success_failure}
\end{figure*}

Across all optimization trajectories, we observe a heavy-tailed distribution of technique usage and success. Figures~\ref{fig:opt_attempts_success_failure} and~\ref{fig:opt_success_counts} summarize the total number of optimization attempts per technique, stacked by success versus failure, and the number of successful applications per technique.

The attempt distribution is dominated by a small number of broadly applicable techniques, while many techniques are explored only rarely. This reflects an optimization process where general-purpose transformations are used as first-order probes across many kernels, and more specialized transformations are invoked selectively.

Successful applications are concentrated in techniques such as SIMD operations, grid size optimization, instruction-level parallelism, block size adaptation, work-per-thread increase, register pressure reduction, fast-math, and thread coarsening. These optimizations are local, broadly applicable, and frequently improve occupancy, instruction throughput, or launch overhead.

Conversely, many high-frequency techniques also exhibit substantial failure mass, indicating that applying common heuristics without profiling-guided state awareness often leads to regressions. This motivates conditioning technique selection on profiling-derived signals rather than relying on uniform or frequency-based application. In summary, these results suggest a two-tier optimization strategy: first, use broadly applicable, low-cost transformations as probes; second, apply more structured transformations, such as memory-layout changes or reduction strategy changes, selectively when profiling indicates the relevant bottleneck.

\subsection{Hardware-Specific Optimizations and Limitations}

In the current evaluation, the agent primarily discovers algebraic simplifications, fusion opportunities, and memory-locality improvements. The generated kernels do leverage Tensor Cores and layout transformations for MMA. While we don’t observe Hopper-specific warp specialization, optimized kernels use SW specialization (producer-consumer, cooperative loading, warp-scoped specialization).

\section{Ablation Studies}

\subsection{Performance Learning Rate}
To understand the impact of {\sys}' \KB on code optimization performance, we conduct a study by limiting the ability of the ICRL flow to learn from experience effectively. As described in Section~\ref{sec:opt-dist}, it is critical to develop and explore a wide range of optimization strategies when optimizing a diverse codebase. However, this diversity relies on {\sys}'s ability to mutate its \KB by modifying and expanding entries.

\begin{figure}[!h]
    \centering
    \includegraphics[width=\columnwidth]{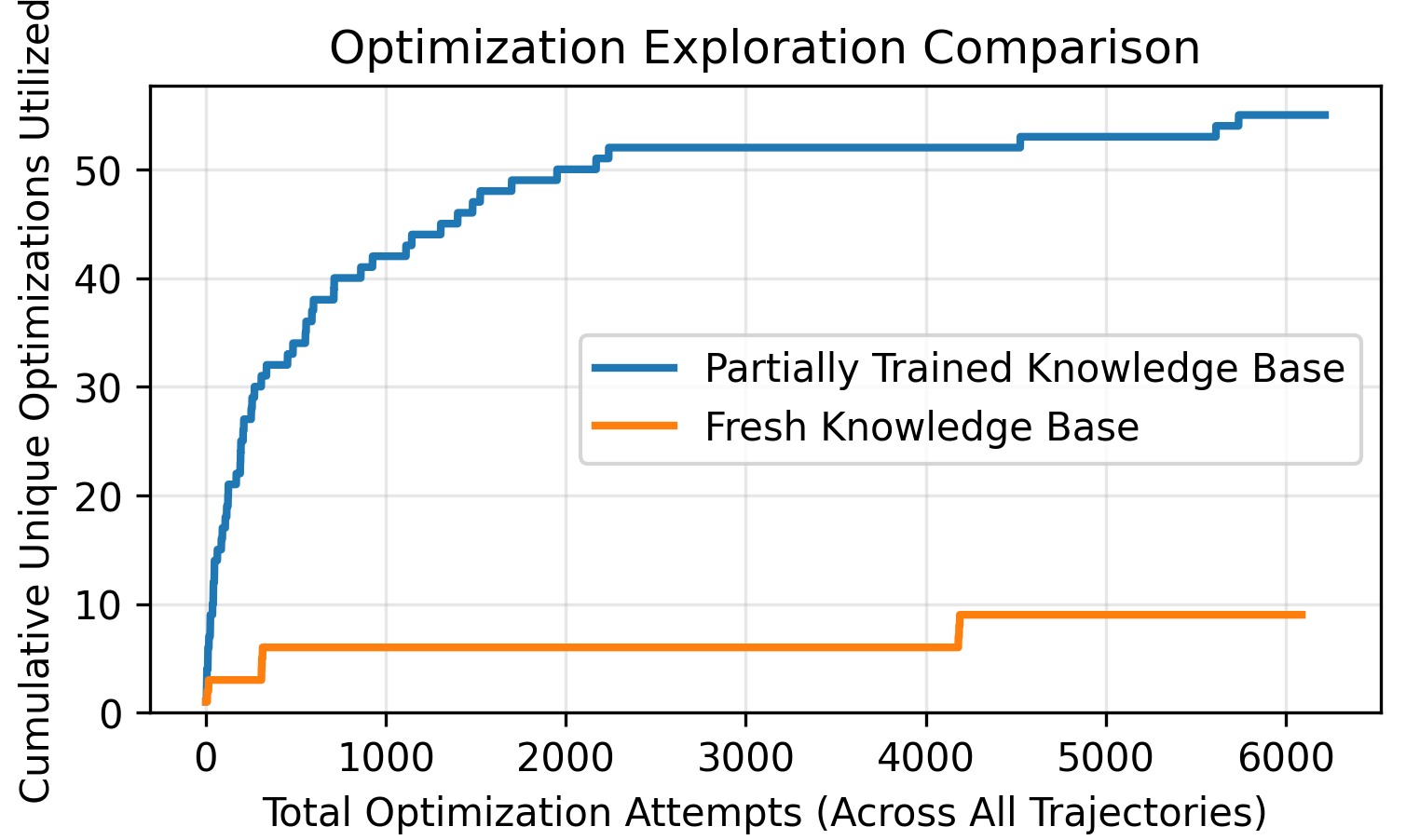}
    \caption{Discovery and application of new optimizations as optimizations are attempted when learning with a pretrained and empty \KB. }
    \label{fig:learning_rate}
\end{figure}

\begin{figure}
    \centering
    \includegraphics[width=\columnwidth]{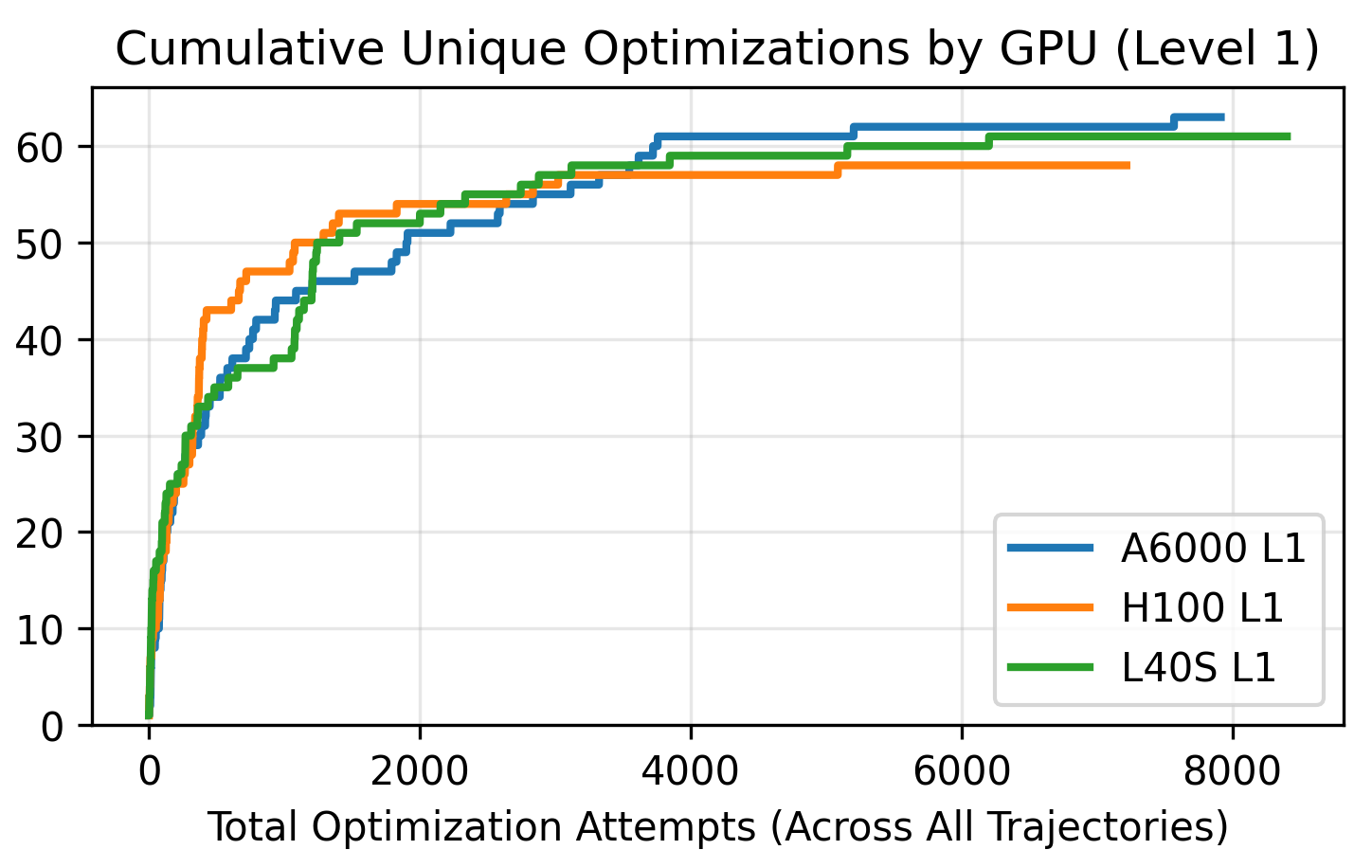}
    \caption{Discovery and application of new optimizations as optimizations are attempted when reusing a \KB trained on A6000 on other GPUs.}
    \label{fig:knowledge_base-learning_rate}
\end{figure}

First, we study the impact of developing a \KB from scratch, compared to iterating upon a partially-trained data structure. The comparison between Figure~\ref{fig:learning_rate} and Figure ~\ref{fig:knowledge_base-learning_rate} shows the rate at which optimizations are applied while optimizing the KernelBench Level 1 dataset. As seen in the left plot, although the first round to construct a \KB is expensive, future optimization passes can benefit from much faster coverage of optimizations. Furthermore, these generated databases can be reused across scenarios;  The right plot demonstrates how a \KB trained on an A6000 GPU can be used for optimization runs across different GPU platforms.

To understand the impact of the long-term memory (\emph{Knowledgebase}), we isolate the contribution of profiling versus memory reuse by comparing against a \texttt{no\_mem\_agent} that has access to full Nsight Compute profiling but operates with an empty knowledge base and no state-conditioned reuse. \texttt{no\_mem\_agent} underperforms our full system, achieving 1.67x slower results. Thus, profiling information alone provides meaningful but limited gains, whereas persistent knowledge reuse is necessary to transfer previously successful optimization strategies across kernels. Ablations show that profiling feedback alone is necessary but not sufficient: the strongest gains arise from the interaction between structured profiling signals and a persistent, state-aware knowledge base.

\subsection{Hyperparameter Analysis} 
\begin{figure}[t]
    \centering
    \includegraphics[width=1.1\columnwidth]{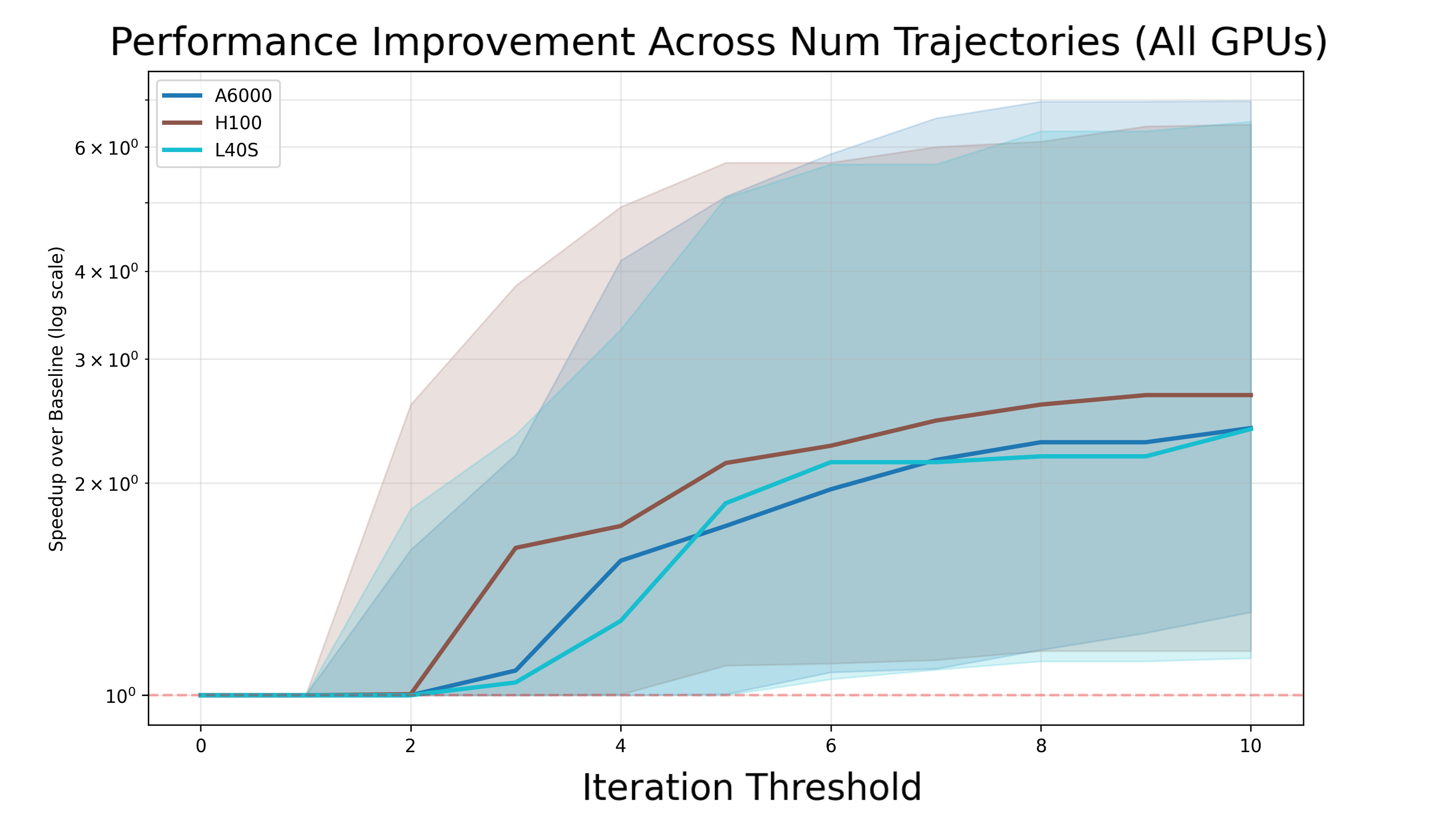}
    \caption{Performance improvement across the number of trajectories. The shaded region shows the inter-quartile range (IQR) of optimized kernels.}
    \label{fig:Num-Traj}
\end{figure}

\begin{figure}[!h]
    \centering
    \includegraphics[width=1.05\columnwidth]{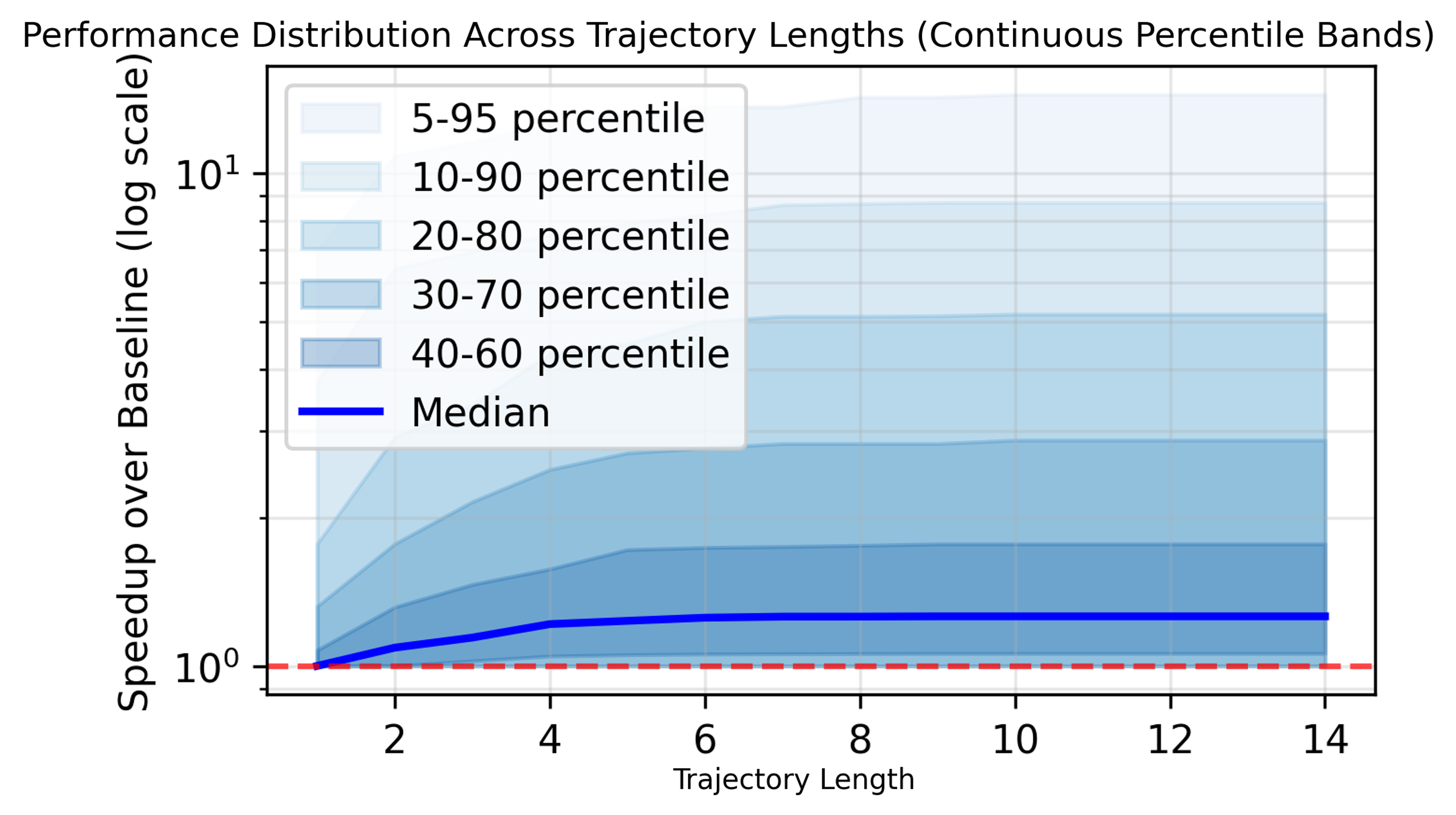}
    \caption{Box plot of performance improvement across trajectory lengths.}
    \label{fig:Traj_Length}
\end{figure}

The key hyperparameters of {\sys} are trajectory length and number of trajectories, corresponding to search depth and breadth, respectively. In Figure~\ref{fig:Num-Traj}, we analyze the impact of search breadth. We observe diminishing returns for increasing trajectory count beyond 8, particularly for median and top 25$^{th}$ kernels. However, we do see additional benefits for the lower 25$^{th}$ percentile kernels.  On the other hand, Figure~\ref{fig:Traj_Length} highlights the diminishing returns of search depth beyond 4 optimizations, as exhausting relevant optimizations to the kernel. An interesting finding, however, is that kernels with high potential for speedup continue to have marginal benefits from additional optimization sequences up to 8 consecutive optimizations, opening up hyperparameters for tuning based on LLM inference cost and performance targets.

\subsection{Profiling Fidelity Analysis}
To understand the impact of profiling fidelity, we conducted ablation studies against an agent that accesses only cycle-level runtime feedback instead of detailed NCU profiles. The cycle-only agent underperforms approaches that use NCU (1.22x speedup over PyTorch on Level-2 compared to 1.57x with NCU data), indicating that scalar latency alone is insufficient to infer why a kernel is slow or which optimization direction to optimize next.

\begin{figure}[t]
  \centering
  \includegraphics[width=\linewidth]{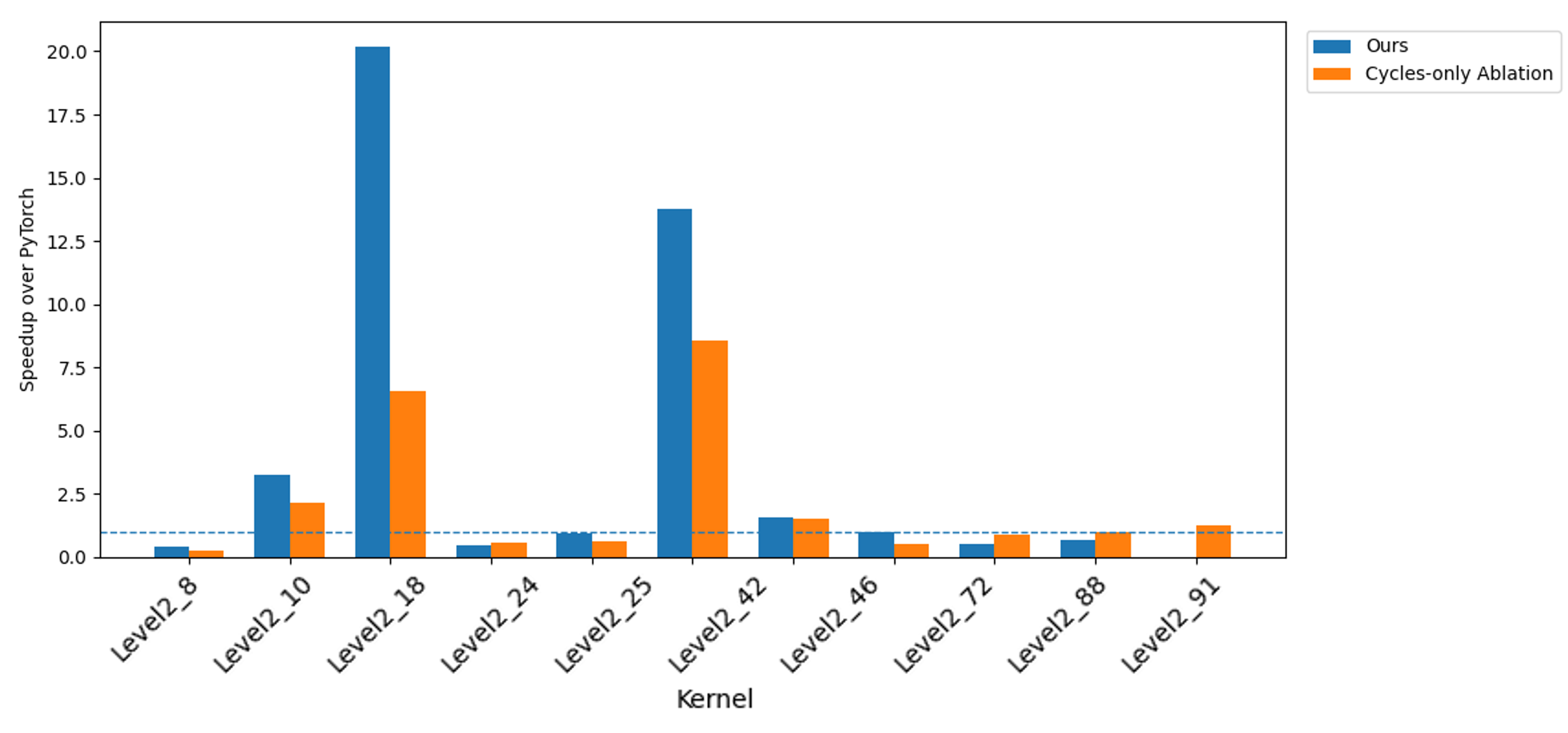}
  \caption{Speedup over PyTorch comparing our full approach against a cycles-only ablation. Removing non-cycle signals degrades performance across most kernels, indicating that additional feedback beyond raw cycle counts is necessary for effective optimization.}
  \label{fig:ours_vs_ablation_cycles_only}
\end{figure}

\subsection{Comparison Against Minimal Agent}
Finally, we compare against a minimal agent. At each iteration, this agent directly takes in CUDA code and NCU profiling data and outputs optimized code. When running 10 trajectories of length 10, this agent requires 2.4x tokens compared to \sys. We observe this is due to two primary causes: 

\begin{enumerate}
    \item Due to not having access to a \emph{Knowledgebase} and a guided reasoning process, this agent must devote more tokens up-front for reasoning about the optimizations
    \item The minimal agent requires more retrievals for correctness compared to \sys.
\end{enumerate}

Compared to \sys, this minimal agent has 0.379x performance improvement per token. Furthermore, \sys achieves better performance in 71\% of cases compared to the simplified agent.

%% file: mlsys2025style/07_conclusion.tex
\section{Conclusions \& Future Work}

In summary, {\sys} demonstrates the potential of In-Context Reinforcement Learning for CUDA optimization, with a geometric mean performance speedup of 1.32x on Kernelbench Level 1 and 2 problems compared to the Pytorch baseline. The results show that LLMs can dynamically adapt optimization strategies at inference time, a capability achieved by learning from a history of successes and failures stored within a novel, compact, domain-specific long-term memory solution.

In addition to providing an agentic code optimization workflow that can achieve performance improvements over prior work, we also provide insights on the characteristics of code optimization workflows. We cover the impacts of optimization diversity, learning ability, and search hyperparameters. Moreover, we observe the impact of different optimization classes, including complex optimization sequences.

Building on the foundation established by {\sys}, our future work focuses on developing more sophisticated solutions to target system performance. To improve Knowledgebase management and reduce storage overheads and bias towards early entries, future directions will explore strategies such as randomized sampling and periodic updates of the Knowledgebase.

To optimize for quality, cost, and latency of LLM calls, we plan to leverage model heterogeneity, a technique that utilizes smaller, more efficient models for simpler agents and more powerful models for more complex agents in the workflow. A critical challenge in code optimization is the problem of phase ordering: the effectiveness of a transformation often depends on the sequence in which it is applied.

Furthermore, {\sys} gradients provide a mechanism for rapid iteration and deployment of customized agents that respond to diverse user needs. With this structure, LLMs can learn not only from immediate interaction traces but also from a growing database of structured experience. While a single agent's database is not useful for updating model parameters, a database aggregated from the full spectrum of agentic systems deployed can be used to train and refine the model itself. This would allow short-term, in-context adaptations to be distilled into long-term, parameter-level improvements, turning accumulated experience into durable capability.

%% file: mlsys2025style/appendix.tex
\clearpage
\onecolumn
\section{Appendix}
\label{sec:appendix}
\subsection{Representative Fused Kernel Launch (KernelBench Level~2 Q18)}

We provide a detailed kernel-level analysis of KernelBench Level~2 Q18, where our method achieves a 20.17$\times$ speedup over the PyTorch baseline. The dominant contributor is an algebraic optimization: after a sequence of reductions, the tensor has shape $(\text{batch\_size}, 1)$, yet \texttt{logsumexp} is applied twice along that dimension. Since $\mathrm{logsumexp}(x, \text{dim}=1) = x$ when the dimension size is one, these operations are algebraically redundant and can be removed exactly without approximation.

Additionally, the generated CUDA implementations further improve performance through kernel fusion (directly producing the final scalar output without materializing an intermediate vector), improved instruction-level parallelism via unrolled accumulation, and efficient block-level reduction using warp shuffles instead of global atomics. These optimizations reduce global memory traffic and kernel launch overhead while preserving correctness.

\begin{lstlisting}[style=cppstyle]
// Warp-level reduction using shuffle
__inline__ __device__ float warp_reduce_sum(float val) {
    unsigned mask = 0xFFFFFFFFu;
    for (int offset = WARP_SIZE / 2; offset > 0; offset >>= 1) {
        val += __shfl_down_sync(mask, val, offset);
    }
    return val;
}

// Block-level reduction built on top of warp reduction
__inline__ __device__ float block_reduce_sum(float val) {
    __shared__ float warp_sums[32]; // supports up to 1024 threads/block
    int lane = threadIdx.x & (WARP_SIZE - 1);
    int warp_id = threadIdx.x >> 5;

    val = warp_reduce_sum(val);
    if (lane == 0) warp_sums[warp_id] = val;
    __syncthreads();

    float out = 0.0f;
    if (warp_id == 0) {
        const int num_warps = (blockDim.x + WARP_SIZE - 1) / WARP_SIZE;
        out = (lane < num_warps) ? warp_sums[lane] : 0.0f;
        out = warp_reduce_sum(out);
    }
    return out;
}

__global__ void model_forward_kernel(__half* __restrict__ output,
                                     const __half* __restrict__ input,
                                     const __half* __restrict__ weight,
                                     const __half* __restrict__ bias,
                                     int batch_size, int in_features, int out_features) {
    const int b = blockIdx.x;
    if (b >= batch_size) return;

    // Stage a tile of the input vector into shared memory for reuse across out_features
    __shared__ float s_x[TILE_K];

    const __half* x_base = input + static_cast<size_t>(b) * in_features;

    // Pre-accumulate the bias terms across the subset of out_features handled by this thread
    float local_bias_sum = 0.0f;
    const int o_iters = (out_features + blockDim.x - 1) / blockDim.x;
#pragma unroll
    for (int it = 0; it < o_iters; ++it) {
        const int o = it * blockDim.x + threadIdx.x;
        if (o < out_features) {
            local_bias_sum += __half2float(bias[o]);
        }
    }

    // Accumulate the dot products across tiles of in_features
    float thread_accum = 0.0f;

    for (int k0 = 0; k0 < in_features; k0 += TILE_K) {
        const int remaining = in_features - k0;
        const int tile = (TILE_K < remaining) ? TILE_K : remaining;

        // Cooperative load of the input tile into shared memory
        for (int t = threadIdx.x; t < tile; t += blockDim.x) {
            s_x[t] = __half2float(x_base[k0 + t]);
        }
        __syncthreads();

        // For this input tile, accumulate contributions for this thread's assigned out_features
#pragma unroll
        for (int it = 0; it < o_iters; ++it) {
            const int o = it * blockDim.x + threadIdx.x;
            if (o < out_features) {
                const __half* __restrict__ w_row = weight + static_cast<size_t>(o) * in_features + k0;

                // Multiple independent accumulators to increase ILP and reduce dependency chains
                float a0 = 0.0f, a1 = 0.0f, a2 = 0.0f, a3 = 0.0f;
                float a4 = 0.0f, a5 = 0.0f, a6 = 0.0f, a7 = 0.0f;

                int t = 0;

                // Main unrolled loop processes UNROLL_T elements per iteration
#pragma unroll
                for (; t + (UNROLL_T - 1) < tile; t += UNROLL_T) {
                    // Load inputs from shared memory
                    float x0 = s_x[t + 0];
                    float x1 = s_x[t + 1];
                    float x2 = s_x[t + 2];
                    float x3 = s_x[t + 3];
                    float x4 = s_x[t + 4];
                    float x5 = s_x[t + 5];
                    float x6 = s_x[t + 6];
                    float x7 = s_x[t + 7];

                    // Convert weights from half to float and accumulate with FMA
                    a0 = fmaf(x0, __half2float(w_row[t + 0]), a0);
                    a1 = fmaf(x1, __half2float(w_row[t + 1]), a1);
                    a2 = fmaf(x2, __half2float(w_row[t + 2]), a2);
                    a3 = fmaf(x3, __half2float(w_row[t + 3]), a3);
                    a4 = fmaf(x4, __half2float(w_row[t + 4]), a4);
                    a5 = fmaf(x5, __half2float(w_row[t + 5]), a5);
                    a6 = fmaf(x6, __half2float(w_row[t + 6]), a6);
                    a7 = fmaf(x7, __half2float(w_row[t + 7]), a7);
                }

                // Tail handling for remaining elements (< UNROLL_T)
                float tail = 0.0f;
#pragma unroll
                for (; t < tile; ++t) {
                    tail = fmaf(s_x[t], __half2float(w_row[t]), tail);
                }

                // Combine partial accumulators
                float partial = (((a0 + a1) + (a2 + a3)) + ((a4 + a5) + (a6 + a7))) + tail;

                thread_accum += partial;
            }
        }
        __syncthreads();
    }

    // Add bias sum after finishing all dot-product tiles
    thread_accum += local_bias_sum;

    // Reduce across the block to produce the final scalar for this batch element
    float block_sum = block_reduce_sum(thread_accum);
    if (threadIdx.x == 0) {
        output[b] = __float2half_rn(block_sum);
    }
}

void launch_gpu_implementation(void* output, void* input, void* weight, void* bias,
                               int batch_size, int in_features, int out_features) {
    if (batch_size <= 0 || in_features <= 0 || out_features <= 0) {
        return;
    }

    // One block per batch element to expose ILP across out_features inside the block.
    // Use a power-of-two block size for efficient reductions.
    const int threads_per_block = 256;
    const int blocks = batch_size; // block per batch sample

    model_forward_kernel<<<blocks, threads_per_block>>>(
        static_cast<__half*>(output),
        static_cast<const __half*>(input),
        static_cast<const __half*>(weight),
        static_cast<const __half*>(bias),
        batch_size, in_features, out_features
    );

    cudaDeviceSynchronize();
}

\end{lstlisting}

\subsection{Representative Tensor Core Optimizations (KernelBench Level~2 063 )}
The implementation realizes an FP16$\times$FP16$\rightarrow$FP32 GEMM using NVIDIA WMMA, where each warp computes a $16\times16$ output tile via $\texttt{mma\_sync}$ and four warps form a $32\times32$ block tile. The $K$ dimension is partitioned across $\texttt{grid.z}$ (split-$K$), so each block accumulates over a disjoint $K$ slice. Full tiles are loaded directly from global memory into WMMA fragments, while boundary tiles are packed with zero padding into per-warp shared-memory buffers to satisfy fragment layout constraints.

Weights stored as $[N\times K]$ row-major are consumed as column-major $\texttt{matrix\_b}$ fragments through pointer and leading-dimension adjustments, avoiding explicit transposition. Partial FP32 accumulations from each split-$K$ slice are written to a global workspace using $\texttt{atomicAdd}$, followed by a separate epilogue kernel that applies bias, ReLU, scaling, and FP16 casting. The design explicitly targets Tensor Core utilization with mixed precision, combines direct-load and packed execution paths for correctness across arbitrary dimensions, isolates per-warp shared-memory regions to avoid cross-warp synchronization, and employs split-$K$ parallelism to increase concurrency at the cost of atomic reduction overhead.

\begin{lstlisting}[style=cppstyle]
#include <cuda_fp16.h>
#include <mma.h>
#include <cuda_runtime.h>
#include <stdint.h>
#include <stddef.h>

using namespace nvcuda;

// Constants and tiling parameters
#define WARP_SIZE 32
#define MMA_M 16
#define MMA_N 16
#define MMA_K 16

// Warp tile mapping: 2x2 tiles per block -> 4 warps
#define TILES_PER_BLOCK_M 2
#define TILES_PER_BLOCK_N 2
#define WARPS_PER_BLOCK (TILES_PER_BLOCK_M * TILES_PER_BLOCK_N)

#define BLOCK_ROWS (TILES_PER_BLOCK_M * MMA_M) // 32
#define BLOCK_COLS (TILES_PER_BLOCK_N * MMA_N) // 32

#define THREADS_PER_BLOCK (WARP_SIZE * WARPS_PER_BLOCK)

// Shared memory sizing per warp: accumulator (float 16x16) + A pack (half 16x16) + B pack (half 16x16)
#define ACCUM_ELEMENTS (MMA_M * MMA_N)                  // 256
#define ACCUM_BYTES    (ACCUM_ELEMENTS * sizeof(float)) // 1024
#define PACK_BYTES     (ACCUM_ELEMENTS * sizeof(__half))// 512

// Align up to 16 bytes
#define ALIGN_UP_16(x) (((x) + 15) & ~((size_t)15))
#define PER_WARP_SHARED_BYTES ALIGN_UP_16((size_t)(ACCUM_BYTES + 2 * PACK_BYTES))

// Simple min/max macros for device code
#define MIN(a,b) (( (a) < (b) ) ? (a) : (b))
#define MAX(a,b) (( (a) > (b) ) ? (a) : (b))

// Helper: compute one 16x16 tile for a specified K-slice [k_start, k_start + k_count)
__device__ __forceinline__ void wmma_tile_16x16_splitk(
    const __half* __restrict__ A_tile_base, int lda,                  // lda = K
    const __half* __restrict__ B_colmajor_base, int ldb,              // ldb = K (weights as [N x K] row-major, loaded as col-major)
    int m_eff, int n_eff,
    int k_start, int k_count,                                         // K-slice control
    float* __restrict__ accum_out,                                    // per-warp 16x16 float tile
    __half* __restrict__ a_pack, __half* __restrict__ b_pack)         // per-warp 16x16 half pack buffers
{
    const int lane_id = threadIdx.x % WARP_SIZE;

    wmma::fragment<wmma::accumulator, MMA_M, MMA_N, MMA_K, float> c_frag;
    wmma::fill_fragment(c_frag, 0.0f);

    int processed = 0;
    while (processed < k_count) {
        const int kk = processed;
        const int k_frag = MIN(MMA_K, k_count - kk);
        const bool full_tile = (m_eff == MMA_M) && (n_eff == MMA_N) && (k_frag == MMA_K);

        if (full_tile) {
            // Fast path: direct global loads
            wmma::fragment<wmma::matrix_a, MMA_M, MMA_N, MMA_K, __half, wmma::row_major> a_frag;
            // Load B^T as col-major to consume weights stored as [N x K] row-major
            wmma::fragment<wmma::matrix_b, MMA_M, MMA_N, MMA_K, __half, wmma::col_major> b_frag;
            // A: row-major, advance by k_start + kk
            wmma::load_matrix_sync(a_frag, A_tile_base + (k_start + kk), lda);
            // B: treat original weight as [N x K] row-major; to obtain B^T (KxN) in col-major, use base pointer at (col=c_col, row=k_start+kk) with ldb=K.
            wmma::load_matrix_sync(b_frag, B_colmajor_base + (k_start + kk), ldb);
            wmma::mma_sync(c_frag, a_frag, b_frag, c_frag);
        } else {
            // Pack with zero padding for partial M/N or K-tail
            for (int idx = lane_id; idx < ACCUM_ELEMENTS; idx += WARP_SIZE) {
                a_pack[idx] = __float2half(0.0f);
                b_pack[idx] = __float2half(0.0f);
            }
            __syncwarp();

            // Pack A (row-major): rows [0..m_eff-1], cols [0..k_frag-1]
            for (int idx = lane_id; idx < ACCUM_ELEMENTS; idx += WARP_SIZE) {
                const int ii = idx / MMA_N; // row in 16x16
                const int jj = idx % MMA_N; // col in 16x16
                if (ii < m_eff && jj < k_frag) {
                    a_pack[ii * MMA_N + jj] = A_tile_base[ii * lda + (k_start + kk + jj)];
                }
            }

            // Pack B into COL-MAJOR buffer:
            // desired element = weight[(c_col + jj) * K + (k_start + kk + ii)]
            // store into b_pack col-major with ld=16: ii + jj * 16
            for (int idx = lane_id; idx < ACCUM_ELEMENTS; idx += WARP_SIZE) {
                const int ii = idx % MMA_K;        // row along K-frag
                const int jj = idx / MMA_K;        // col along N-frag
                if (ii < k_frag && jj < n_eff) {
                    b_pack[ii + jj * MMA_K] = B_colmajor_base[(k_start + kk + ii) + jj * ldb];
                }
            }
            __syncwarp();

            wmma::fragment<wmma::matrix_a, MMA_M, MMA_N, MMA_K, __half, wmma::row_major> a_frag_pack;
            wmma::fragment<wmma::matrix_b, MMA_M, MMA_N, MMA_K, __half, wmma::col_major> b_frag_pack;
            wmma::load_matrix_sync(a_frag_pack, a_pack, MMA_N);          // row-major with ld=16
            wmma::load_matrix_sync(b_frag_pack, b_pack, MMA_K);          // col-major with ld=16
            wmma::mma_sync(c_frag, a_frag_pack, b_frag_pack, c_frag);
        }

        processed += k_frag;
    }

    // Store accumulator to per-warp float tile buffer (row-major 16x16)
    wmma::store_matrix_sync(accum_out, c_frag, MMA_N, wmma::mem_row_major);
    __syncwarp();
}

// Kernel 1: Split-K partial GEMM accumulation into a global float workspace using atomicAdd
__global__ void splitk_gemm_accumulate_kernel(
    const __half* __restrict__ A,     // [M x K], row-major
    const __half* __restrict__ B,     // weight as [N x K], row-major
    float* __restrict__ workspace,    // [M x N], row-major (float accumulators), pre-zeroed
    int M, int N, int K,
    int K_per_slice)
{
    const int warp_id = threadIdx.x / WARP_SIZE;

    // Map each warp to a unique 16x16 tile inside a 2x2 tile group per block
    const int warp_m = warp_id / TILES_PER_BLOCK_N; // 0..(TILES_PER_BLOCK_M-1)
    const int warp_n = warp_id % TILES_PER_BLOCK_N; // 0..(TILES_PER_BLOCK_N-1)

    const int c_row_base = blockIdx.y * BLOCK_ROWS + warp_m * MMA_M;
    const int c_col_base = blockIdx.x * BLOCK_COLS + warp_n * MMA_N;

    // Check bounds and compute effective tile sizes
    const int m_eff = MAX(0, MIN(MMA_M, M - c_row_base));
    const int n_eff = MAX(0, MIN(MMA_N, N - c_col_base));
    if (m_eff <= 0 || n_eff <= 0) {
        return;
    }

    // K-slice for this block in grid.z
    const int slice_id = blockIdx.z;
    const int k_start = slice_id * K_per_slice;
    const int k_count = MAX(0, MIN(K_per_slice, K - k_start));
    if (k_count <= 0) return;

    // Dynamic shared memory layout (per-warp independent regions)
    extern __shared__ unsigned char shared_bytes[];
    // Ensure 16-byte alignment of our base pointer
    uintptr_t base_addr = reinterpret_cast<uintptr_t>(shared_bytes);
    base_addr = (base_addr + 15u) & ~((uintptr_t)15u);
    unsigned char* shmem_aligned = reinterpret_cast<unsigned char*>(base_addr);

    unsigned char* warp_shmem_base = shmem_aligned + warp_id * PER_WARP_SHARED_BYTES;

    float* accum_out = reinterpret_cast<float*>(warp_shmem_base);
    __half* a_pack   = reinterpret_cast<__half*>(warp_shmem_base + ACCUM_BYTES);
    __half* b_pack   = reinterpret_cast<__half*>(warp_shmem_base + ACCUM_BYTES + PACK_BYTES);

    // Base pointers for this tile
    const __half* A_tile_base = A + c_row_base * K;       // row-major MxK
    // Treat weight B as [N x K] row-major; to load B^T as col-major fragments,
    // set base pointer at the start of column block j=c_col_base: &B[j*K]
    const __half* B_tile_col  = B + c_col_base * K;       // base for col-major load with ldb=K

    // Compute partial GEMM over [k_start, k_start + k_count)
    wmma_tile_16x16_splitk(
        A_tile_base, K,
        B_tile_col,  K,
        m_eff, n_eff,
        k_start, k_count,
        accum_out, a_pack, b_pack
    );

    // Atomically accumulate partial results into global float workspace
    const int lane_id = threadIdx.x % WARP_SIZE;
    const int total_out_elems = m_eff * n_eff;
    for (int idx = lane_id; idx < total_out_elems; idx += WARP_SIZE) {
        const int i = idx / n_eff; // 0..m_eff-1
        const int j = idx % n_eff; // 0..n_eff-1
        const int global_row = c_row_base + i;
        const int global_col = c_col_base + j;
        const float val = accum_out[i * MMA_N + j];
        atomicAdd(&workspace[global_row * N + global_col], val);
    }
}

// Kernel 2: Epilogue applying bias + ReLU + division and cast to __half
__global__ void epilogue_bias_relu_div_kernel(
    const float* __restrict__ workspace, // [M x N], row-major (float accumulators)
    const __half* __restrict__ bias,     // [N]
    __half* __restrict__ C,              // [M x N], row-major
    float divisor,
    int M, int N)
{
    const float inv_div = 1.0f / divisor;
    const int tid = blockIdx.x * blockDim.x + threadIdx.x;
    const int total_elems = M * N;

    for (int idx = tid; idx < total_elems; idx += gridDim.x * blockDim.x) {
        const int col = idx % N;
        float val = workspace[idx];
        val += __half2float(bias[col]);
        val = fmaxf(val, 0.0f) * inv_div;
        C[idx] = __float2half_rn(val);
    }
}

// Host launcher: split-K two-pass implementation
void launch_gpu_implementation(void* output, void* input, void* weight, void* bias, float divisor,
                               int batch_size, int in_features, int out_features) {
    // Problem sizes
    const int M = batch_size;
    const int K = in_features;
    const int N = out_features;

    // Tile grid covering MxN by 32x32 blocks (4 warps per block)
    dim3 grid_xy(
        (N + BLOCK_COLS - 1) / BLOCK_COLS,
        (M + BLOCK_ROWS - 1) / BLOCK_ROWS
    );

    // Heuristic for split-K: target ~256 K per slice, up to 8 slices
    int split_k_slices = 1;
    if (K >= 256) {
        const int target_per_slice = 256;
        split_k_slices = (K + target_per_slice - 1) / target_per_slice;
        if (split_k_slices > 8) split_k_slices = 8;
        if (split_k_slices < 1) split_k_slices = 1;
    }

    // Final 3D grid with split-K in z
    dim3 grid(grid_xy.x, grid_xy.y, split_k_slices);
    dim3 block(THREADS_PER_BLOCK);

    // Dynamic shared memory: per-warp region times number of warps
    size_t shmem_size = WARPS_PER_BLOCK * PER_WARP_SHARED_BYTES;

    // Workspace for partial accumulation (float)
    float* workspace = nullptr;
    size_t workspace_bytes = (size_t)M * (size_t)N * sizeof(float);
    cudaMalloc(&workspace, workspace_bytes);
    cudaMemset(workspace, 0, workspace_bytes);

    // Launch partial GEMM with split-K accumulation
    const int K_per_slice = (K + split_k_slices - 1) / split_k_slices;
    splitk_gemm_accumulate_kernel<<<grid, block, shmem_size>>>(
        static_cast<const __half*>(input),    // A: [M x K]
        static_cast<const __half*>(weight),   // B: [N x K] row-major (consumed as B^T)
        workspace,                            // float accumulators
        M, N, K,
        K_per_slice
    );

    // Epilogue kernel: bias + ReLU + divide -> output half
    const int threads = 256;
    const int total_elems = M * N;
    const int blocks = (total_elems + threads - 1) / threads;
    epilogue_bias_relu_div_kernel<<<blocks, threads>>>(
        workspace,
        static_cast<const __half*>(bias),
        static_cast<__half*>(output),
        divisor,
        M, N
    );

    // Cleanup
    cudaFree(workspace);
    cudaDeviceSynchronize();
}
\end{lstlisting}

\subsection{Level~3 Full Model Example (SqueezeNetFireModule)}
We provide a representative sample of \sys applied to more complex Level3 problems from KernelBench; this example includes a full implementation a SqueezeNet Fire Module. \sys optimizes over multiple kernels and kernel launches, demonstrating the capacity to optimize over more complex CUDA codebases, and achieving a speedup of 1.2$\times$ over the PyTorch baseline.

This implementation speeds up conv/pool/linear layers by assigning one CUDA block per output element and parallelizing the reduction dimension cooperatively across threads, then combining partial results with shuffle-based warp reductions and a lightweight block reduction via a per-warp shared-memory staging buffer. It reduces memory overhead by fusing bias and (when applicable) ReLU into the conv and linear kernels, uses \texttt{\_\_restrict\_\_} and force-inlined helpers to aid compilation, and routes most input/weight/bias reads through the read-only cache path (\texttt{\_\_ldg}). Computation is performed with FP16 storage and FP32 accumulation, avoiding extra reshape work by treating the post-pool tensor as a flattened vector by pointer aliasing.

\begin{lstlisting}[style=cppstyle]
#include <cuda_runtime.h>
#include <cuda_fp16.h>
#include <stdint.h>
#include <cstdio>
#include <cmath>

// Simple CUDA error checker (debugging aid)
#ifndef NDEBUG
#define CUDA_CHECK(x) do { cudaError_t err = (x); if (err != cudaSuccess) { \
    fprintf(stderr, "CUDA error %s at %s:%d\n", cudaGetErrorString(err), __FILE__, __LINE__); abort(); } } while (0)
#else
#define CUDA_CHECK(x) x
#endif

// Constants
#define WARP_SIZE 32
#ifndef WPT
#define WPT 4
#endif

inline int div_up_int(int a, int b) { return (a + b - 1) / b; }
inline int div_up_int64(int64_t a, int64_t b) { return (int)((a + b - 1) / b); }

// Helper: accumulate dot product of half2 pairs into float
__device__ __forceinline__ void accumulate_half2_pair(const __half2 a, const __half2 b, float &acc) {
    float2 af = __half22float2(a);
    float2 bf = __half22float2(b);
    acc += af.x * bf.x + af.y * bf.y;
}

// Read-only cached load helpers
template <typename T>
__device__ __forceinline__ T ro_load(const T* ptr) {
#if __CUDA_ARCH__ >= 350
    return __ldg(ptr);
#else
    return *ptr;
#endif
}

// Vectorized load of two consecutive halfs as half2 with alignment check.
// Fallback to two scalar loads if unaligned.
__device__ __forceinline__ __half2 ro_load2_aligned(const half* base, int idx) {
    const half* p = base + idx;
    if ((((uintptr_t)p) & 0x3) == 0) {
#if __CUDA_ARCH__ >= 350
        return __ldg(reinterpret_cast<const __half2*>(p));
#else
        return *reinterpret_cast<const __half2*>(p);
#endif
    } else {
        return __halves2half2(ro_load(base + idx), ro_load(base + idx + 1));
    }
}

// Aligned store of two halfs via half2 if aligned and within bounds; fallback otherwise.
__device__ __forceinline__ void store2_aligned(half* base, int idx, __half2 v, int64_t limit) {
    half* p = base + idx;
    if ((((uintptr_t)p) & 0x3) == 0 && (int64_t)(idx + 1) < limit) {
        *reinterpret_cast<__half2*>(p) = v;
    } else {
        float2 vf = __half22float2(v);
        base[idx] = __float2half_rn(vf.x);
        if ((int64_t)(idx + 1) < limit) {
            base[idx + 1] = __float2half_rn(vf.y);
        }
    }
}

// Warp- and block-level reductions (sum and max)
__device__ __forceinline__ float warpReduceSum(float val) {
    unsigned mask = 0xFFFFFFFFu;
    for (int offset = WARP_SIZE / 2; offset > 0; offset >>= 1) {
        val += __shfl_down_sync(mask, val, offset);
    }
    return val;
}
__device__ __forceinline__ float warpReduceMax(float val) {
    unsigned mask = 0xFFFFFFFFu;
    for (int offset = WARP_SIZE / 2; offset > 0; offset >>= 1) {
        val = fmaxf(val, __shfl_down_sync(mask, val, offset));
    }
    return val;
}

__device__ __forceinline__ float blockReduceSum(float val) {
    static __shared__ float smem[32]; // supports up to 32 warps per block
    int lane = threadIdx.x & (WARP_SIZE - 1);
    int wid  = threadIdx.x / WARP_SIZE;
    val = warpReduceSum(val);
    if (lane == 0) smem[wid] = val;
    __syncthreads();
    float out = 0.f;
    if (wid == 0) {
        int numWarps = (blockDim.x + WARP_SIZE - 1) / WARP_SIZE;
        out = (lane < numWarps) ? smem[lane] : 0.f;
        out = warpReduceSum(out);
    }
    return out;
}

__device__ __forceinline__ float blockReduceMax(float val) {
    static __shared__ float smem[32]; // supports up to 32 warps per block
    int lane = threadIdx.x & (WARP_SIZE - 1);
    int wid  = threadIdx.x / WARP_SIZE;
    val = warpReduceMax(val);
    if (lane == 0) smem[wid] = val;
    __syncthreads();
    float out = -INFINITY;
    if (wid == 0) {
        int numWarps = (blockDim.x + WARP_SIZE - 1) / WARP_SIZE;
        out = (lane < numWarps) ? smem[lane] : -INFINITY;
        out = warpReduceMax(out);
    }
    return out;
}

// Algorithmic-change optimized kernels: cooperative block reductions over reduction dimensions.

// Conv2d NCHW with bias and ReLU, stride, padding; cooperative reduction across C_in*kH*kW
__global__ void conv2d_nchw_bias_relu_reduce_kernel(
    const half* __restrict__ input,   // [N, C_in, H, W]
    const half* __restrict__ weight,  // [C_out, C_in, kH, kW]
    const half* __restrict__ bias,    // [C_out]
    half* __restrict__ output,        // [N, C_out, OH, OW]
    int N, int C_in, int H, int W,
    int C_out, int kH, int kW,
    int stride_h, int stride_w,
    int pad_h, int pad_w,
    int OH, int OW
) {
    int tidx = blockIdx.x;
    if (tidx >= N * C_out * OH * OW) return;

    // Map linear index to (n, oc, oh, ow)
    int ow = tidx % OW; tidx /= OW;
    int oh = tidx % OH; tidx /= OH;
    int oc = tidx % C_out;
    int n  = tidx / C_out;

    const int in_n_stride = C_in * H * W;
    const int in_c_stride = H * W;
    const int w_oc_stride = C_in * kH * kW;
    const int w_ic_stride = kH * kW;

    int in_h0 = oh * stride_h - pad_h;
    int in_w0 = ow * stride_w - pad_w;

    float acc = 0.f;

    const int R = C_in * kH * kW; // reduction length
    for (int r = threadIdx.x; r < R; r += blockDim.x) {
        int ic  = r / (kH * kW);
        int rem = r % (kH * kW);
        int kh  = rem / kW;
        int kw  = rem % kW;

        int h_in = in_h0 + kh;
        int w_in = in_w0 + kw;
        if ((unsigned)h_in < (unsigned)H && (unsigned)w_in < (unsigned)W) {
            int in_idx = n * in_n_stride + ic * in_c_stride + h_in * W + w_in;
            int w_idx  = oc * w_oc_stride + ic * w_ic_stride + kh * kW + kw;
            half hin = ro_load(&input[in_idx]);
            half hw  = ro_load(&weight[w_idx]);
            acc += __half2float(hin) * __half2float(hw);
        }
    }

    float sum = blockReduceSum(acc);
    if (threadIdx.x == 0) {
        sum += __half2float(ro_load(&bias[oc]));
        sum = sum < 0.f ? 0.f : sum;
        int out_idx = n * (C_out * OH * OW) + oc * (OH * OW) + oh * OW + ow;
        output[out_idx] = __float2half_rn(sum);
    }
}

// MaxPool2d NCHW without padding: cooperative reduction across kH*kW window
__global__ void maxpool2d_nchw_reduce_kernel(
    const half* __restrict__ input,  // [N, C, H, W]
    half* __restrict__ output,       // [N, C, OH, OW]
    int N, int C, int H, int W,
    int kH, int kW,
    int stride_h, int stride_w,
    int OH, int OW
) {
    int tidx = blockIdx.x;
    if (tidx >= N * C * OH * OW) return;

    int ow = tidx % OW; tidx /= OW;
    int oh = tidx % OH; tidx /= OH;
    int c  = tidx % C;
    int n  = tidx / C;

    int h_start = oh * stride_h;
    int w_start = ow * stride_w;

    float local_max = -INFINITY;

    const int in_n_stride = C * H * W;
    const int in_c_stride = H * W;

    const int R = kH * kW;
    for (int r = threadIdx.x; r < R; r += blockDim.x) {
        int kh = r / kW;
        int kw = r % kW;

        int h_in = h_start + kh;
        int w_in = w_start + kw;
        if ((unsigned)h_in < (unsigned)H && (unsigned)w_in < (unsigned)W) {
            int row_base = n * in_n_stride + c * in_c_stride + h_in * W;
            float v = __half2float(ro_load(&input[row_base + w_in]));
            local_max = fmaxf(local_max, v);
        }
    }

    float max_val = blockReduceMax(local_max);
    if (threadIdx.x == 0) {
        int out_idx = n * (C * OH * OW) + c * (OH * OW) + oh * OW + ow;
        output[out_idx] = __float2half_rn(max_val);
    }
}

// Linear: Y = ReLU(X * W^T + b); cooperative reduction across K
__global__ void linear_gemm_bias_relu_reduce_kernel(
    const half* __restrict__ X,      // [N, K]
    const half* __restrict__ W,      // [M, K]
    const half* __restrict__ bias,   // [M]
    half* __restrict__ Y,            // [N, M]
    int N, int M, int K
) {
    int tidx = blockIdx.x;
    if (tidx >= N * M) return;

    int m = tidx % M;
    int n = tidx / M;

    const half* x_ptr = X + n * K;
    const half* w_ptr = W + m * K;

    float acc = 0.f;
    for (int k = threadIdx.x; k < K; k += blockDim.x) {
        acc += __half2float(ro_load(&x_ptr[k])) * __half2float(ro_load(&w_ptr[k]));
    }

    float sum = blockReduceSum(acc);
    if (threadIdx.x == 0) {
        sum += __half2float(ro_load(&bias[m]));
        sum = sum < 0.f ? 0.f : sum;
        Y[n * M + m] = __float2half_rn(sum);
    }
}

// Linear: Y = X * W^T + b; cooperative reduction across K (no activation)
__global__ void linear_gemm_bias_reduce_kernel(
    const half* __restrict__ X,      // [N, K]
    const half* __restrict__ W,      // [M, K]
    const half* __restrict__ bias,   // [M]
    half* __restrict__ Y,            // [N, M]
    int N, int M, int K
) {
    int tidx = blockIdx.x;
    if (tidx >= N * M) return;

    int m = tidx % M;
    int n = tidx / M;

    const half* x_ptr = X + n * K;
    const half* w_ptr = W + m * K;

    float acc = 0.f;
    for (int k = threadIdx.x; k < K; k += blockDim.x) {
        acc += __half2float(ro_load(&x_ptr[k])) * __half2float(ro_load(&w_ptr[k]));
    }

    float sum = blockReduceSum(acc);
    if (threadIdx.x == 0) {
        sum += __half2float(ro_load(&bias[m]));
        Y[n * M + m] = __float2half_rn(sum);
    }
}

// Public entry point used by the harness
void launch_gpu_implementation(
    void* output,
    const void* input,
    const void* conv1_weight,
    const void* conv1_bias,
    const void* conv2_weight,
    const void* conv2_bias,
    const void* fc1_weight,
    const void* fc1_bias,
    const void* fc2_weight,
    const void* fc2_bias,
    const void* fc3_weight,
    const void* fc3_bias,
    int64_t batch_size,
    int64_t in_channels,
    int64_t in_h,
    int64_t in_w,
    // Conv1 params
    int64_t conv1_out_channels,
    int64_t conv1_kernel_h,
    int64_t conv1_kernel_w,
    int64_t conv1_stride_h,
    int64_t conv1_stride_w,
    int64_t conv1_pad_h,
    int64_t conv1_pad_w,
    // Pool params
    int64_t pool_kernel_h,
    int64_t pool_kernel_w,
    int64_t pool_stride_h,
    int64_t pool_stride_w,
    // Conv2 params
    int64_t conv2_out_channels,
    int64_t conv2_kernel_h,
    int64_t conv2_kernel_w,
    int64_t conv2_stride_h,
    int64_t conv2_stride_w,
    int64_t conv2_pad_h,
    int64_t conv2_pad_w,
    // Linear params
    int64_t fc1_in_features,
    int64_t fc1_out_features,
    int64_t fc2_out_features,
    int64_t fc3_out_features
) {
    // Cast inputs to half pointers
    const half* x_in  = static_cast<const half*>(input);
    const half* w1    = static_cast<const half*>(conv1_weight);
    const half* b1    = static_cast<const half*>(conv1_bias);
    const half* w2    = static_cast<const half*>(conv2_weight);
    const half* b2    = static_cast<const half*>(conv2_bias);
    const half* wfc1  = static_cast<const half*>(fc1_weight);
    const half* bfc1  = static_cast<const half*>(fc1_bias);
    const half* wfc2  = static_cast<const half*>(fc2_weight);
    const half* bfc2  = static_cast<const half*>(fc2_bias);
    const half* wfc3  = static_cast<const half*>(fc3_weight);
    const half* bfc3  = static_cast<const half*>(fc3_bias);
    half* y_out       = static_cast<half*>(output);

    // Shapes and params
    const int N  = static_cast<int>(batch_size);
    const int C0 = static_cast<int>(in_channels);
    const int H0 = static_cast<int>(in_h);
    const int W0 = static_cast<int>(in_w);

    const int C1  = static_cast<int>(conv1_out_channels);
    const int K1H = static_cast<int>(conv1_kernel_h);
    const int K1W = static_cast<int>(conv1_kernel_w);
    const int S1H = static_cast<int>(conv1_stride_h);
    const int S1W = static_cast<int>(conv1_stride_w);
    const int P1H = static_cast<int>(conv1_pad_h);
    const int P1W = static_cast<int>(conv1_pad_w);

    const int PKH = static_cast<int>(pool_kernel_h);
    const int PKW = static_cast<int>(pool_kernel_w);
    const int PSH = static_cast<int>(pool_stride_h);
    const int PSW = static_cast<int>(pool_stride_w);

    const int C2  = static_cast<int>(conv2_out_channels);
    const int K2H = static_cast<int>(conv2_kernel_h);
    const int K2W = static_cast<int>(conv2_kernel_w);
    const int S2H = static_cast<int>(conv2_stride_h);
    const int S2W = static_cast<int>(conv2_stride_w);
    const int P2H = static_cast<int>(conv2_pad_h);
    const int P2W = static_cast<int>(conv2_pad_w);

    const int FC1_K = static_cast<int>(fc1_in_features);   // Expect 16*5*5 typically
    const int FC1_M = static_cast<int>(fc1_out_features);  // 120
    const int FC2_M = static_cast<int>(fc2_out_features);  // 84
    const int FC3_M = static_cast<int>(fc3_out_features);  // num_classes

    // Derived dims (floor semantics)
    const int H1 = (H0 + 2 * P1H - K1H) / S1H + 1;
    const int W1 = (W0 + 2 * P1W - K1W) / S1W + 1;

    const int H1p = (H1 - PKH) / PSH + 1;
    const int W1p = (W1 - PKW) / PSW + 1;

    const int H2 = (H1p + 2 * P2H - K2H) / S2H + 1;
    const int W2 = (W1p + 2 * P2W - K2W) / S2W + 1;

    const int H2p = (H2 - PKH) / PSH + 1;
    const int W2p = (W2 - PKW) / PSW + 1;

    // Buffers for intermediates
    half *conv1_out = nullptr, *pool1_out = nullptr;
    half *conv2_out = nullptr, *pool2_out = nullptr;
    half *fc1_out   = nullptr, *fc2_out   = nullptr;

    size_t conv1_bytes = static_cast<size_t>(N) * C1 * H1 * W1 * sizeof(half);
    size_t pool1_bytes = static_cast<size_t>(N) * C1 * H1p * W1p * sizeof(half);
    size_t conv2_bytes = static_cast<size_t>(N) * C2 * H2 * W2 * sizeof(half);
    size_t pool2_bytes = static_cast<size_t>(N) * C2 * H2p * W2p * sizeof(half);
    size_t fc1_bytes   = static_cast<size_t>(N) * FC1_M * sizeof(half);
    size_t fc2_bytes   = static_cast<size_t>(N) * FC2_M * sizeof(half);

    CUDA_CHECK(cudaMalloc(&conv1_out, conv1_bytes));
    CUDA_CHECK(cudaMalloc(&pool1_out, pool1_bytes));
    CUDA_CHECK(cudaMalloc(&conv2_out, conv2_bytes));
    CUDA_CHECK(cudaMalloc(&pool2_out, pool2_bytes));
    CUDA_CHECK(cudaMalloc(&fc1_out,   fc1_bytes));
    CUDA_CHECK(cudaMalloc(&fc2_out,   fc2_bytes));

    // Choose cooperative-reduction block size
    const int threads = 128; // good balance for reductions on small K/R; adjust if needed

    // Conv1 + Bias + ReLU (cooperative reduction)
    {
        int total = N * C1 * H1 * W1;
        int blocks = total;
        if (blocks < 1) blocks = 1;
        conv2d_nchw_bias_relu_reduce_kernel<<<blocks, threads>>>(
            x_in, w1, b1, conv1_out,
            N, C0, H0, W0,
            C1, K1H, K1W,
            S1H, S1W, P1H, P1W,
            H1, W1
        );
        CUDA_CHECK(cudaGetLastError());
    }
    // MaxPool1 (reads already-rectified conv1_out) - cooperative max reduction
    {
        int total = N * C1 * H1p * W1p;
        int blocks = total;
        if (blocks < 1) blocks = 1;
        // Note: preserve original PSW special-case logic for first pool (as in source)
        maxpool2d_nchw_reduce_kernel<<<blocks, threads>>>(
            conv1_out, pool1_out,
            N, C1, H1, W1,
            PKH, PKW, PSH, PSH == 0 ? 1 : PSW,
            H1p, W1p
        );
        CUDA_CHECK(cudaGetLastError());
    }

    // Conv2 + Bias + ReLU (cooperative reduction)
    {
        int total = N * C2 * H2 * W2;
        int blocks = total;
        if (blocks < 1) blocks = 1;
        conv2d_nchw_bias_relu_reduce_kernel<<<blocks, threads>>>(
            pool1_out, w2, b2, conv2_out,
            N, C1, H1p, W1p,
            C2, K2H, K2W,
            S2H, S2W, P2H, P2W,
            H2, W2
        );
        CUDA_CHECK(cudaGetLastError());
    }
    // MaxPool2 (reads already-rectified conv2_out) - cooperative max reduction
    {
        int total = N * C2 * H2p * W2p;
        int blocks = total;
        if (blocks < 1) blocks = 1;
        maxpool2d_nchw_reduce_kernel<<<blocks, threads>>>(
            conv2_out, pool2_out,
            N, C2, H2, W2,
            PKH, PKW, PSH, PSW,
            H2p, W2p
        );
        CUDA_CHECK(cudaGetLastError());
    }

    // Flatten pool2_out [N, C2, H2p, W2p] to [N, K_flat] by pointer alias
    const int K_flat = C2 * H2p * W2p;
    (void)K_flat;

    // FC1: Fused GEMM + Bias + ReLU (cooperative reduction across K)
    {
        int total = N * FC1_M;
        int blocks = total;
        if (blocks < 1) blocks = 1;
        linear_gemm_bias_relu_reduce_kernel<<<blocks, threads>>>(
            pool2_out, wfc1, bfc1, fc1_out,
            N, FC1_M, FC1_K
        );
        CUDA_CHECK(cudaGetLastError());
    }

    // FC2: Fused GEMM + Bias + ReLU (cooperative reduction across K)
    {
        int total = N * FC2_M;
        int blocks = total;
        if (blocks < 1) blocks = 1;
        linear_gemm_bias_relu_reduce_kernel<<<blocks, threads>>>(
            fc1_out, wfc2, bfc2, fc2_out,
            N, FC2_M, FC1_M
        );
        CUDA_CHECK(cudaGetLastError());
    }

    // FC3 -> output (no activation) (cooperative reduction across K)
    {
        int total = N * FC3_M;
        int blocks = total;
        if (blocks < 1) blocks = 1;
        linear_gemm_bias_reduce_kernel<<<blocks, threads>>>(
            fc2_out, wfc3, bfc3, y_out,
            N, FC3_M, FC2_M
        );
        CUDA_CHECK(cudaGetLastError());
    }

    // Ensure kernels finish before freeing
    CUDA_CHECK(cudaDeviceSynchronize());

    // Free temporaries
    CUDA_CHECK(cudaFree(conv1_out));
    CUDA_CHECK(cudaFree(pool1_out));
    CUDA_CHECK(cudaFree(conv2_out));
    CUDA_CHECK(cudaFree(pool2_out));
    CUDA_CHECK(cudaFree(fc1_out));
    CUDA_CHECK(cudaFree(fc2_out));
}

\end{lstlisting}
